\documentclass[final,5p,times,twocolumn]{elsarticle}
\usepackage[utf8]{inputenc}

\usepackage{algorithm}
\usepackage{algpseudocode}
\usepackage{subfigure,underoverlap}
\usepackage{multirow}
\usepackage{color}
\usepackage{cancel}
\usepackage{soul}
\usepackage{hyperref}

\title{Time series cluster kernels to exploit informative missingness and incomplete label information}

\author[MLIFT,UNN2]{Karl Øyvind Mikalsen\corref{cor1} }

\cortext[cor1]{Corresponding author at: UiT Machine Learning Group  (machine-learning.uit.no), Department of Physics and Technology, UiT The Arctic University of Norway, NO-9037 Tromsø, Norway}
\address[MLIFT]{Dept. of Physics and Technology, UiT The Arctic University of Norway, NO-9037 Tromsø, Norway}

\ead{karl.o.mikalsen@uit.no}

\author[MLIFT,spain]{Cristina Soguero-Ruiz}
\address[spain]{Dept. of Signal Theory and Comm., Telematics and Computing, Universidad Rey Juan Carlos, Fuenlabrada, Spain}

\author[MLIFT]{Filippo Maria Bianchi}

\author[UNN2,UNN3,UNN4]{Arthur Revhaug}
\address[UNN2]{Dept. of Gastrointestinal Surgery, University Hospital of North Norway (UNN), Tromsø, Norway}
\address[UNN3]{Clinic for Surgery, Cancer and Women's Health, UNN, Tromsø, Norway}
\address[UNN4]{Institute of Clinical Medicine, UiT, Tromsø, Norway}

\author[MLIFT]{Robert Jenssen}

\begin{document}

\begin{frontmatter}

\begin{abstract}
The time series cluster kernel (TCK) provides a powerful tool for analysing multivariate time series subject to missing data.
TCK is designed using an ensemble learning approach in which Bayesian mixture models form the base models. Because of the Bayesian approach, TCK can naturally deal with missing values without resorting to imputation and the ensemble strategy ensures robustness to hyperparameters, making it particularly well suited for unsupervised learning.

However, TCK assumes missing at random and that the underlying missingness mechanism is ignorable, i.e. uninformative, an assumption that does not hold in many real-world applications, such as e.g. medicine.
To overcome this limitation,  we present a kernel capable of exploiting the potentially rich information in the missing values and patterns, as well as the information from the observed data. In our approach,  we create a representation of the missing pattern, which is incorporated into mixed mode mixture models in such a way that the information provided by the missing patterns is effectively exploited. Moreover, we also propose a semi-supervised kernel,  capable of  taking advantage of incomplete label information  to learn more accurate similarities.

Experiments on benchmark data, as well as a real-world case study of patients described by longitudinal electronic health record data who potentially suffer from hospital-acquired infections, demonstrate the effectiveness of the proposed methods. 
\end{abstract}


\end{frontmatter}


\section{Introduction}
\label{sec: intro}

Multivariate time series (MTS) frequently occur in a whole range of practical applications such as medicine, biology, and climate studies, to name a few. A challenge that complicates the analysis is that real-world MTS are often subject to large amounts of missing data. Traditionally, missingness mechanisms have been categorized into missing completely at random (MCAR), missing at random (MAR) and missing not at random (MNAR)~\cite{rubin1976inference}. 
The main difference between these mechanisms consists in whether the missingness is ignorable (MCAR and MAR) or non-ignorable (MNAR)~\cite{rubin1976inference, molenberghs2009incomplete, doi:10.1111/j.1467-9868.2007.00640.x}. In e.g. medicine, non-ignorable missingness can occur when the missing patterns $R$ are related to the disease under study $Y$. In this case, the distribution of the missing patterns for diseased patients is not equal to the corresponding distribution for the control group, i.e. $p(R \: | \: Y = 1) \neq p(R \: | \: Y = 0)$. Hence, the missingness is \textit{informative}~\cite{ allen2003informative,Guo2005,  DBLP:journals/corr/ChePCSL16}. By contrast, uninformative missingness will be referred to as \textit{ignorable} in the remainder of this paper.

Both ignorable and informative missingness occur in real-world data. An example from medicine of ignorable missingness occurs e.g. if a clinician orders lab tests for a patient and the tests are performed, but because of an error the results are not recorded. On the other hand, informative missingness could occur if it is decided to not perform  lab tests because the doctor thinks the patient is in good shape. 
In the latter case, the missing values and patterns potentially contain rich information about the diseases and clinical outcomes for the patient.
Efficient data-driven approaches aiming to extract knowledge, perform predictive modeling, etc.,  must be capable of capturing this information.

Various methods have been proposed to handle missing data in MTS~\cite{schafer2002missing, schafer1997analysis, little2014statistical}.
One simple approach is to create a \textit{complete} dataset by  discarding the time series with missing data. However, this gives unbiased predictions only if the missingness mechanism is MCAR. 
As an alternative, a preprocessing step involving \emph{imputation} of missing values with some estimated value, such as the mean, is common. Other so-called \emph{single imputation} methods exploit machine learning based methods such as multilayer perceptrons, self-organizing maps, k-nearest neighbors, recurrent neural networks and regression-based imputation~\cite{garcia2010pattern, RAHMAN2015198}. 
Alternatively, one can impute missing values using various smoothing and interpolation techniques~\cite{ENGELS2003968, garcia2010pattern}. Among these, a prominent example is the last observation carried forward (LOCF) scheme that imputes the last non-missing value for the following missing values.
Limitations of imputation methods are that they introduce additional bias and they ignore uncertainty associated with the missing values.

\emph{Multiple imputation}~\cite{white2011multiple} resolves this problem, to some extent, by estimating the missing values multiple times and thereby creating multiple complete datasets. Thereafter, e.g. a classifier is trained on all datasets and the results are combined to obtain the final predictions.
However, despite that multiple imputation and other imputation methods can give satisfying results in some scenarios,  these are ad-hoc solutions that lead to a multi-step procedure in which the missing data are handled separately and independently from the rest of the analysis. 
Moreover, the information about which values are actually missing (the missing patterns) is lost, i.e. imputation methods cannot exploit informative missingness. 

Due to the aforementioned limitations, several research efforts have been devoted over the last years to process incomplete time series without relying on imputation~\cite{DBLP:journals/corr/ChePCSL16, bianchi2018time, mikalsen2016learning, pmlr-v56-Lipton16, bianchi2018learning, Marlin:2012:UPD:2110363.2110408, Ghassemi:2015:MTM:2887007.2887070}.
In this regard, powerful kernel methods have been proposed, of which the recently proposed   \emph{time series cluster kernel} (TCK)~\cite{mikalsen2017time} is a prominent example. 
The TCK is designed using an ensemble learning approach in which Bayesian mixture models form the base models. 
An advantage of TCK, compared to imputation methods, is that the missing data are handled automatically and no additional tasks are left to the user. Multiple imputation instead requires a careful selection of the imputation model and other variables are needed to do the imputation~\cite{schafer2002missing},  which  particularly in an unsupervised setting can turn out to be problematic.

A shortcoming of the TCK is that unbiased predictions are only guaranteed for ignorable missingness, i.e. the kernel cannot take advantage of informative missing patterns frequently occurring in medical applications.
To overcome this limitation, in this work, we present a novel time series cluster kernel, TCK$_{IM}$.
In our approach,  we create a representation of the missing patterns using masking, i.e.  we represent the  missing patterns  using binary indicator time series. By doing so, we obtain MTS consisting of both continuous and discrete attributes.  To model these time series, we introduce mixed mode Bayesian mixture models, which can effectively exploit information provided by the missing patterns.

The time series cluster kernels are particularly useful in unsupervised settings. In many practical applications such as e.g. medicine it is not feasible to obtain completely labeled training sets~\cite{MIKALSEN2017105}, but in some cases it is possible to annotate a few samples with labels, i.e. incomplete label information is available. In order to exploit the incomplete label information, we propose a semi-supervised MTS kernel, ssTCK.  In our approach,   we incorporate ideas from information theory to measure similarities between distributions. More specifically, we employ the Kullback-Leibler divergence to assign labels to unlabeled data. 

Experiments on benchmark MTS datasets and a real-world case study of patients suffering from hospital-acquired infections, described by longitudinal electronic health record data, demonstrate the effectiveness of the proposed TCK$_{IM}$ and ssTCK kernels. 

The remainder of this paper is organized as follows. Section~\ref{sec: background} presents background on MTS kernels. The two proposed kernels are described in Section~\ref{sec: tck im} and \ref{sec: sstck}, respectively. Experiments on synthetic and benchmark datasets are presented in Section~\ref{sec: experiments}, whereas the case study is described in Section~\ref{Sec: case study}. Section~\ref{sec: conclusion} concludes the paper.

\section{Multivariate time series kernels to handle missing data}
\label{sec: background}
Kernel methods have been of great importance in machine learning for several decades and have applications in many different fields~\cite{Jenssen2010,camps2009kernel,soguero2016support}.
Within the context of time series, a \textit{kernel} is a similarity measure that also is positive semi-definite~\cite{shawe2004kernel}. 
Once defined, such similarities between pairs of time series may be utilized in a wide range of applications, such as classification or clustering, benefiting from the vast body of work in the field of kernel methods.  
Here we provide an overview of MTS kernels, and describe how they deal with missing data.


The simplest of all kernel functions is the linear kernel, which for two data points represented as vectors, $x$ and $y$, is given by the inner product $\langle x, y \rangle$, possibly plus a constant $c$. 
One can also apply a linear kernel to pairs of MTS once  they are unfolded into vectors. However, by doing so the information that they are MTS and there might be inherent dependencies in time and between  attributes, is then lost. 
Nevertheless, in some cases such a kernel can be efficient, especially if the MTS are short~\cite{chen2013model}.
If the MTS contain missing data, the linear kernel requires a preprocessing step involving e.g. imputation.


The most widely used time series similarity measure is \emph{dynamic time warping} (DTW) \cite{Berndt:1994:UDT:3000850.3000887}, where the similarity is quantified as the alignment cost between the MTS. More specifically, in DTW the time dimension of one or both of the time series is warped to achieve a better alignment. 
Despite the success of DTW in many applications, similarly to many other similarity measures, it is non-metric and therefore cannot non-trivially be used to design a positive semi-definite kernel~\cite{marteau2015recursive}. Hence, it is not suited for kernel methods in its original formulation.
However, because of its popularity there have been attempts to design kernels exploiting the DTW. For example, Cuturi et al. designed a DTW-based kernel using global alignments~\cite{cuturi2007kernel}. 
An efficient version of the global alignment kernel (GAK) is provided in~\cite{cuturi2011fast}. 
The latter has two hyperparameters, namely the kernel bandwidth and the triangular parameter. 
GAK does not naturally deal with missing data and incomplete datasets, and therefore also requires a preprocessing step involving imputation.

Two MTS kernels that can naturally deal with missing data without having to resort to imputation are the  \emph{learned pattern similarity} (LPS)~\cite{baydogan2016time} and TCK.
LPS generalizes the well-known autoregressive models
to local autopatterns using multiple lag values for autocorrelation. These autopatterns are supposed to capture the local dependency structure in the time series and are learned using a tree-based (random forest) learning strategy. 
More specifically, a time series is represented as a matrix of segments. Randomness is injected to the learning process by randomly choosing time segment (column in the matrix) and lag $p$ for each tree in the random forest. A bag-of-words type compressed representation is created from the output of the leaf-nodes for each tree. The final time series representation is created by concatenating the representation obtained from the individual trees, which in turn are used to compute the similarity using a histogram intersection kernel~\cite{barla2003histogram}.

The TCK is based on an ensemble learning approach 
wherein  robustness to hyperparameters is ensured by joining the clustering results of many Gaussian mixture models (GMM) to form the final kernel. Hence, no critical hyperparameters have to be tuned by the user, and the TCK can be learned in an unsupervised manner. 
To ensure robustness to sparsely sampled data, the GMMs that are the base models in the ensemble,  are extended using informative prior distributions such that the missing data is explicitly dealt with.
More specifically, the TCK matrix is built by fitting GMMs to the set of MTS for a range of number of mixture components. The idea is that by generating partitions at different resolutions, one can capture both the local and global structure of the data. 
Moreover, to capture diversity in the data, randomness is injected by for each resolution (number of components) estimating the mixture parameters for a range of random initializations and randomly chosen hyperparameters. In addition, each GMM sees a random subset of attributes and segments in the MTS. 
The posterior distributions for each mixture component are then used to build the TCK matrix by taking the inner product between all pairs of posterior distributions.
Eventually, given an ensemble of GMMs, the TCK is created in an additive way by using the fact that the sum of kernels is also a kernel.

Despite that LPS and TCK kernels share many properties, the way missing data are dealt with is very different. In LPS,  the missing data handling abilities of decision trees are exploited. Along with ensemble methods, fuzzy approaches and support vector solutions, decision trees can be categorized as \emph{machine learning approaches for handling missing data}~\cite{garcia2010pattern}, i.e. the missing data are handled naturally by the machine learning algorithm. 
One can also argue that the way missing data are dealt with in the TCK belongs to this category, since an ensemble approach is exploited. However, it can also be categorized as a \emph{likelihood-based approach} since the underlying models in the ensemble are Gaussian mixture models. In the likelihood-based approaches, the full, incomplete dataset is analysed using maximum likelihood (or maximum a posteriori, equivalently), typically in combination with the expectation-maximization (EM) algorithm~\cite{schafer2002missing, little2014statistical}. These approaches assume that the missingness is ignorable.

\section{Time series cluster kernel to exploit informative missingness}
\label{sec: tck im}
In this section, we present the novel time series cluster kernel, TCK$_{IM}$, which is capable of exploiting informative missingness.  

A key component in the time series cluster kernel framework is ensemble learning, in which the basic idea consists in combining a collection of many base models into a composite model. A good such composite model will have  statistical, computational and representational advantages  such as lower variance, lower sensitivity to local optima and is capable of representing a broader span functions (increased expressiveness), respectively, compared to the individual base models~\cite{Dietterich2000}. Key to achieve this is \emph{diversity} and \emph{accuracy}~\citep{hansen1990neural}, i.e. the base models cannot make the same errors on new test data and have to perform better than random guessing. This can be done by integrating multiple outcomes of the same  (weak) base model as it is trained under different, often randomly chosen, settings (parameters, initialization, subsampling, etc.) to ensure diversity~\citep{vega2011survey}.

In the TCK$_{IM}$ kernel, the base model is a mixed mode Bayesian mixture model. Next, we provide the details of this model.

\subsection*{Notation}
The following notation is used.
A  multivariate time series (MTS) $X$ is defined as a (finite) combination of univariate time series (UTS),
$
X = \{ x_v \in \mathbb{R}^T \: | \: v = 1,2,\dots,V\},
$
where each attribute, $x_v$, is a UTS of length $T$. The number of UTS, $V$, is the \textit{dimension} of $X$. The length $T$ of the UTS $x_v$ is also the length of the MTS $X$. Hence,  a $V$--dimensional MTS, $X$, of length $T$ can be represented as a matrix in $\mathbb{R}^{V \times T}$. 
Given a dataset of $N$ MTS, we denote $X^{(n)}$ the $n$-th MTS. 
An incompletely observed MTS is described by the pair $U^{(n)} = (X^{(n)}, R^{(n)})$, where $R^{(n)}$ is a binary MTS 
with entry $r_v^{(n)}(t) = 0$ if the realization $x_v^{(n)}(t)$ is missing and $r_v^{(n)}(t) = 1$ if it is observed.

\subsection*{Mixed mode mixture model}

Assume that a MTS $ U = (X, R)$ is generated from two modes. $X$ is a V-variate real-valued MTS ($ X \in \mathbb{R}^{V \times T}$), whereas $R$ is a V-variate binary MTS ($ R \in \{0,1\}^{V \times T}$).    
Further, we assume that $U$ is generated from a finite mixture density,
\begin{equation}
    p(U \: | \: \Phi, \Theta ) = \sum_{g=1}^G \theta_g f( U \: | \: \phi_g),
\end{equation}
where $G$ is the number of components, 
$f$ is the density of the components parametrized by $\Phi = (\phi_1, \dots, \phi_G)$,
and $\Theta = (\theta_1, \dots, \theta_g) $ are the mixing coefficients, $0 \leq \theta_G \leq 1$ and $ \sum_{g=1}^G \theta_g = 1$. 

Now, introduce a latent random variable $ Z $, represented as a $G$-dimensional one-hot vector $Z = (Z_1,\dots,Z_G)$, whose marginal distribution is given by 
$
p(Z \: | \: \Theta ) = \prod_{g=1}^G  \theta_g^{Z_g}.
$
The unobserved variable $Z$ records the membership of $U$ and therefore $Z_g = 1$ if $U$ belongs to component $g$ and $Z_g = 0 $ otherwise.
Hence,
$
    p(U \: | Z, \: \Phi)  = \prod_{g=1}^G f( U \: | \: \phi_g)^{Z_g},
$
and therefore it follows that
\begin{align}
    p(U, Z \: | \:  \Phi, \Theta) 
     =p(U \: | Z, \: \Phi)  p(Z \: | \: \Theta ) = \prod_{g=1}^G \left[  f(U \: | \:   \phi_g) \theta_g \right]^{Z_g}
\end{align}
$U = (X, R) $ consists of two modalities $X$ and $R$. We now naively assume that
\begin{align}
    f(U \: | \:   \phi_g) = f(X \: | \: R, \mu_g, \Sigma_g ) f( R \: | \: \beta_g), 
\end{align}
where $ f(X \: | \: R, \mu_g, \Sigma_g  ) $ is a density function given by
\begin{equation}  \label{eq: diag gmm density}
    f(X \: | \: R, \mu_g, \Sigma_g  ) =  \prod_{v=1}^V \prod_{t=1}^T  \mathcal{N} (x_v(t) \: | \: \mu_{gv}(t), \sigma_{gv})^{r_v(t) },
\end{equation}
and $f( R \: | \: \beta_g) $ is a probability mass given by
\begin{equation}  \label{eq: BMM prob mass}
    f( R \: | \: \beta_g) =  \prod_{v=1}^V \prod_{t=1}^T \beta_{gvt}^{r_v(t)} (1 - \beta_{gvt})^{1-r_v(t)}.
\end{equation}
The parameters of each component are $\phi_g = (\mu_g, \Sigma_g, \beta_g) $, where $\mu_g = \{ \mu_{gv} \in  \mathbb{R}^T \: | \: v = 1,...,V\}$ is a time-dependent mean ($\mu_{gv}$ is a UTS of length $T$),
$\Sigma_g = diag\{\sigma_{g1}^2,...,\sigma_{gV}^2\}$ is
a time-constant diagonal covariance matrix in which $\sigma_{gv}^2$ is the variance of attribute $v$,
and  $\beta_{gvt} \in [0,1]$ are the parameters of the Bernoulli mixture model~\eqref{eq: BMM prob mass}.
The idea is that even though the missingness mechanism is ignored in $ f(X \: | \: R, \mu_g, \Sigma_g  ) $, which is only computed over the observed data, the Bernoulli term $f( R \: | \: \beta_g) $ will capture information from the missing patterns.

The conditional probability of $Z$ given $U$, can be found using Bayes' theorem,
\begin{align} \label{eq: p(z|x) posterior}
\pi_{g} &\equiv P(Z_g = 1 \: | \: U,  \Phi, \Theta )  \nonumber \\
= & \footnotesize{\frac{ \theta_g \prod\limits_{v=1}^V \prod\limits_{t=1}^T  \left[ \mathcal{N} (x_v(t) \: | \: \mu_{gv}(t), \sigma_{gv}) \beta_{gvt}\right]^{r_v(t)}  (1 - \beta_{gvt})^{1-r_v(t)} }{\sum\limits_{g=1}^G \theta_g \prod\limits_{v=1}^V \prod\limits_{t=1}^T  \left[ \mathcal{N} (x_v(t) \: | \: \mu_{gv}(t), \sigma_{gv}) \beta_{gvt}\right]^{r_v(t)}  (1 - \beta_{gvt})^{1-r_v(t)}}}.
\end{align}

Similarly to~\cite{mikalsen2017time}, we introduce a Bayesian extension and put informative priors over the parameters of the normal distribution, which enforces smoothness over time and that clusters containing few time series, to have parameters similar to the mean and covariance computed over the whole dataset.
A kernel-based Gaussian prior is defined for the mean,
$
P(\mu_{gv}) = \mathcal{N} \left(\mu_{gv} \: | \: m_{v}, \: S_{v}\right).
$
$m_{v}$ are the empirical means and the prior covariance matrices, $S_{v}$, are defined as
$
S_{v} = s_{v} \mathcal{K},
$
where $s_{v}$ are empirical standard deviations and $\mathcal{K}$ is a kernel matrix, whose elements are
$
\mathcal{K}_{tt'} = b_0 \exp (-a_0(t-t')^2), \quad t, \, t' = 1,\dots,T.
$
$a_0$, $b_0$ are user-defined hyperparameters.
 An inverse Gamma distribution prior is put on the standard deviation $\sigma_{gv}$,
$
P(\sigma_{gv}) \propto \sigma_{gv}^{-N_0} \exp \left(- \frac{N_0 s_v}{2 \sigma_{gv}^2} \right),
$
where $N_0$ is a user-defined hyperparameter. We denote $\Omega = \{ a_0, b_0, N_0\}$ the set of hyperparameters.

Then, given a dataset $\{U^{(n)}\}_{n=1}^N$, the parameters $\{ \Phi, \Theta \}$ can be estimated using maximum a posteriori expectation maximization (MAP-EM)~\cite{dempster1977maximum, mclachlan2007algorithm}. This leads to  
Algorithm \ref{alg:algorithm 2}.
%
\begin{algorithm}[!t]
\small
\caption{MAP-EM for mixed mode mixture model}
\label{alg:algorithm 2}
\begin{algorithmic}[1]
\Require Dataset $\{U^{(n)} = (X^{(n)}, R^{(n)} )  \}_{n=1}^N$, hyperparameters $\Omega$ and number of mixtures $G$.
\State Initialize the parameters $\Theta =  (\theta_1, \dots, \theta_G) $ and $\Phi = \{\mu_g, \sigma_g, \beta_g\}_ {g=1}^G$.
\State E-step. For each MTS $U^{(n)}$, evaluate the posterior probabilities using Eq.~\eqref{eq: p(z|x) posterior} with the current parameter estimates.
\State M-step. Update parameters using the current posteriors
\begin{align*}
\theta_g &= N^{-1} \textstyle \sum_{n=1}^{N} \pi_{g}^{(n)} 
\\
\sigma_{gv}^2 &=  \frac{ N_0 s^2_{v} + \sum_{n=1}^N \sum_{t=1}^T r^{(n)}_v(t) \; \pi^{(n)}_{g} \big(x^{(n)}_v(t) - \mu_{gv}(t)\big)^2 }{N_0 +  \sum_{n=1}^N \sum_{t=1}^T r^{(n)}_v(t) \; \pi_{g}^{(n)}}
\\
\mu_{gv} &= 
 \frac{ S^{-1}_{v} m_{v} +  \sigma^{-2}_{gv} \textstyle \sum_{n=1}^N  \pi^{(n)}_{g} \text{diag}(r^{(n)}_{v} ) \: x^{(n)}_v }{S^{-1}_{v} + \sigma^{-2}_{gv} \textstyle \sum_{n=1}^N  \pi^{(n)}_{g} \text{diag}(r^{(n)}_{v})} \\
  \beta_{gvt} &= (\textstyle \sum_{n=1}^{N} \pi_{g}^{(n)})^{-1} \textstyle \sum_{n=1}^{N} \pi_{g}^{(n)}  r_v^{(n)}(t)
\end{align*} 
\State Repeat step 2-3 until convergence.
\Ensure Posteriors $ \Pi^{(n)} \equiv \left( \pi_1^{(n)},\dots,\pi_G^{(n)} \right)^T $ and  parameter estimates $\Theta$ and $\Phi$.
\end{algorithmic}
\end{algorithm}

\subsection{Forming the kernel}
We now explain how the mixed mode mixture model is used to form the TCK$_{IM}$ kernel.

We use the mixed mode Bayesian mixture model as the base model in an ensemble approach.
To ensure diversity, we vary the number of components for the base models by sampling from  a set of integers $ \mathcal{I}_C = \{I,\dots, I+C\}$. For each number of components, we apply $Q$ different  random initial conditions and hyperparameters. We let $\mathcal{Q} = \{ q = (q_1,q_2) \: | \: q_1=1,\dots Q, \: q_2 \in \mathcal{I}_C \} $ be the index set keeping track of initial conditions and hyperparameters ($q_1$), and the number of components ($q_2$).
Each base model $q$ is trained on a random subset of MTS $\{(X^{(n)}, R^{(n)} )  \}_{n \in \eta(q)}$. Moreover, for each  $q$, we select random subsets of variables $\mathcal{V}(q)$  as well as random time segments $\mathcal{T}(q)$.

The inner products of the normalized posterior distributions from each mixture component are then added up to build the TCK$_{IM}$ kernel matrix. Note that, in addition to introducing novel base models to account for informative missingness,  we also modify the kernel  by normalizing the vectors of posteriors to have unit length in the $l_2$-norm. This provides an additional regularization that may increase the generalization capability of the learned model.
The details of the method are presented in Algorithm~\ref{alg:algorithm}. The kernel for MTS not available during training can be evaluated according to Algorithm~\ref{alg:algorithm out of sample}.

\begin{algorithm}[t!]
\small
\caption{Time series cluster kernel. Training phase.}
\label{alg:algorithm}
\begin{algorithmic}[1]
\Require Training set of MTS $\{(X^{(n)}, R^{(n)} )  \}_{n=1}^N$ , $Q$ initializations, set of integers $\mathcal{I}_C $ controlling number of components for each base model.
\State Initialize kernel matrix $K = 0_{N \times N}  $.
\For{$q \in \mathcal{Q}$}
\State Compute posteriors $ \Pi^{(n)}(q) \equiv ( \pi_1^{(n)},\dots,\pi_{q_2}^{(n)} )^T $, by fitting a mixed mode mixture model with $q_2$ clusters to the dataset and by randomly selecting:
\begin{itemize}
\item[i.] hyperparameters $\Omega(q) $,
\item[ii.] a time segment $ \mathcal{T}(q)  $ of length {\small $T_{min} \leq  |\mathcal{T}(q)| \: \leq \: T_{max}$ } to extract from each $X^{(n)}$ and $R^{(n)}$,
\item[iv.] a subset of attributes $\mathcal{V}(q)$, with cardinality {\small$V_{min} \leq |\mathcal{V}(q)| \leq V_{max}$}, to extract from each $X^{(n)}$ and $R^{(n)}$,
\item[vi.] a subset of MTS, $\eta(q) $, with {\small$N_{min} \leq |\eta(q)| \leq N$},
\item[vii.] initialization of the mixture parameters $ \Theta(q) $ and $\Phi(q)$.
\end{itemize}
\State Update kernel matrix, $K_{nm} = K_{nm} + \frac{\Pi^{(n)}(q)^T \Pi^{(m)}(q)}{ \| \Pi^{(n)}(q) \| \cdot \| \Pi^{(m)}(q) \| } $.
\EndFor
\Ensure $K$ kernel matrix, time segments $\mathcal{T}(q)  $, subsets of attributes $\mathcal{V}(q)$, subsets of MTS $\eta(q)$, parameters $ \Theta(q)$, $\Phi(q)$  and posteriors $\Pi^{(n)}(q) $.
\end{algorithmic}
\end{algorithm}

\begin{algorithm}[t]
\small
\caption{Time series cluster kernel. Test phase.}
\label{alg:algorithm out of sample}
\begin{algorithmic}[1]
  \Require Test set $\big \{ X^{*(m)} \big \}_{m=1}^M$, time segments $\mathcal{T}(q)  $ subsets of attributes $\mathcal{V}(q)$, $\mathcal{V}_R(q)$, subsets of MTS $\eta(q)$, parameters $ \Theta(q)$, $\Phi(q)$  and posteriors $\Pi^{(n)}(q) $.
  \State Initialize kernel matrix $K^* = 0_{N \times M} $. 
  \For{$q \in \mathcal{Q}$}
  \State Compute posteriors $\Pi^{*(m)}(q) $, $m=1,\dots,M$ using the mixture parameters $ \Theta(q)$, $\Phi(q)$.
  \State Update kernel matrix, $K^*_{nm} = K^*_{nm} + \frac{\Pi^{(n)}(q)^T \Pi^{*(m)}(q)}{ \| \Pi^{(n)}(q) \| \cdot \| \Pi^{*(m)}(q) \| } $.
  \EndFor
  \Ensure $K^*$ test kernel matrix.
\end{algorithmic}
\end{algorithm}

\section{Semi-supervised time series cluster kernel}
\label{sec: sstck}
This section presents a semi-supervised MTS kernel, ssTCK, capable of exploiting incomplete label information. In ssTCK, the base mixture models are learned exactly in the same way as in TCK or TCK$_{IM}$. I.e. if there is no missing data, or the missingness is ignorable, the base models will be the Bayesian GMMs. Conversely, if the missingness is informative, the base models are the mixed mode Bayesian mixture models presented in the previous section.
Both approaches will associate each MTS $X^{(n)}$ with a $q_2$-dimensional posterior $ \Pi^{(n)} \equiv \left( \pi_1^{(n)},\dots,\pi_{q_2}^{(n)} \right)^T $, where $\pi_g^{(n)}$ represents the probability that the MTS belongs to component $g$ and $q_2$ is the total number of components in the base mixture model. 

In ssTCK,  label information is incorporated in an intermediate processing step in which the posteriors $ \Pi^{(n)}$ are transformed,  before the transformed posteriors are sent into Algorithm~\ref{alg:algorithm} or \ref{alg:algorithm out of sample}.
More precisely, the transformation consists in mapping the posterior for the mixture components to a class "posterior" (probability), i.e. we seek to find a function $\mathcal{M}: \: [0,1]^{q_2} \to [0,1]^{N_c}$,
$ 
 \Pi^{(n)} \overset{\mathcal{M}}{\longrightarrow}  \tilde{\Pi}^{(n)}.
$
Hence, we want to exploit the incomplete label information to find a transformation that merges the $q_2$ components of the mixture model into $N_c$ clusters, where $N_c$ is the number of classes.

The mapping $\mathcal{M}$ can be thought of as a (soft) $N_c$-class classifier, and hence there could be many possible ways of learning $\mathcal{M}$. However, choosing a too flexible classifier for this purpose leads to an increased risk of overfitting and could also unnecessarily increase the algorithmic complexity. For these reasons, we restrict ourselves to searching for a linear transformation 
\begin{equation} \label{eq: posterior trans}
 \mathcal{M}( \Pi^{(n)}) = W^T  \Pi^{(n)}, \quad W \in [0,1]^{q_2 \times N_c}.
\end{equation}
Since the $N_c$-dimensional output $ \tilde{\Pi}^{(n)} =  \mathcal{M}( \Pi^{(n)})$ should  represent a probability distribution, we add the constraint $ \sum_{i=1}^{N_c} W_{ji} = 1$, $j = 1,\dots,q_2 $.

A natural first step is to first assume that the label information is complete and look at the corresponding supervised kernel.
In the following two subsections, we describe our proposed methods for learning the transformation $\mathcal{M}$ in supervised and semi-supervised settings, respectively.

\subsection{Supervised time series cluster kernel (sTCK)}
\textit{Supervised setting.} Each base mixture model consists of $q_2$ components, and we assume that the number of components is greater or equal to the number of classes $N_c$. Further, assume that each MTS $X^{(n)}$ in the training set is associated with a $N_c$--dimensional one-hot vector $y^{(n)}$, which represents its label. Hence, the labels of the training set can be represented via a matrix $Y \in \{0,1\}^{N \times N_c}$, where $N$ is the number of MTS in the training set.  

\begin{algorithm}[t]
\small
\caption{Supervised posterior transformation}
\label{alg:algorithm supervised posterior transformation}
\begin{algorithmic}[1]
  \Require Posteriors $\{\Pi^{(n)}\}_ {n=1}^N $ from mixture models consisting of $q_2$ components and labels $\{y^{(n)}\}_ {n=1}^N$,
  \For{$i = 1, \dots, q_2$, $j = 1, \dots, N_c$}
  \State Compute $W_{ij} = \frac{\sum_{n=1}^N y_j^{(n)} \pi_i^{(n)} }{\sum_{n=1}^N y_j^{(n)}}$.
  \State $W_{ij} = \frac{W_{ij}}{\sum_{j=1}^{N_c} W_{ij} }$.
  \EndFor
  \State Transform training and test posteriors via   $\tilde{\Pi} = W^T \Pi   $
  \Ensure Transformed posteriors $\tilde{\Pi}^{(n)} $ 
\end{algorithmic}
\end{algorithm}

We approach this problem by considering one component at the time. For a given component $g$, the task is to associate it with a class. One natural way to do this is to identify all members of component $g$ and then simply count how many times each label occur. To account for class imbalance, one can then divide each count by the number of MTS in the corresponding class. One possible option would then be to assign the component to the class with the largest normalized count. However, by doing so, one is not accounting for uncertainty/disagreement within the component. Hence, a more elegant alternative is to simply use the normalized counts as the weights in the matrix $W$. Additionally, one has to account for that each MTS can simultaneously belong to several components, i.e. each MTS $X^{(n)}$ has a only soft membership to the component $g$, determined by the value $\pi_g^{(n)}$. This can be done using $\Pi^{(n)}$ as weights in the first step. This procedure is summarized in Algorithm~\ref{alg:algorithm supervised posterior transformation}.

\subsection{Semi-supervised time series cluster kernel (ssTCK)}
\textit{Setting.} Assume that the labels $\{y^{(n)}\}_ {n=1}^L$, $L < N$, are known and  $\{y^{(n)}\}_ {n=L+1}^N$ are unknown.

In this setting, if one naively tries to apply Algorithm~\ref{alg:algorithm supervised posterior transformation} based on only the labeled part of the dataset, one  ends up dividing by 0s. The reason is that some of the components in the mixture model will contain only unlabeled MTS (the soft label analogy is that the probability that any of the labeled MTS belong to that particular component is zero or very close to zero). Hence, we need a way to assign labels to the components that do not contain any labeled MTS.

Note that each component is described by a probability distribution. A natural measure of  dissimilarity between probability distributions is the Kullback-Leibler (KL) divergence~\cite{kullback1951information}. Moreover, since the components are described by parametric distributions, the KL divergence has a simple closed-form expression. The KL divergence between two components, $i$ and $j$, in our Bayesian GMM is given by
\begin{align}
    D_{KL}( f^{(i)} \: \| \: f^{(j)} ) =& \frac{1}{2}\bigg( \sum_{v=1}^V \sum_{t=1}^T \sigma_{iv}^2 \sigma_{jv}^{-2} + \sigma_{jv}^{-2}(\mu_{jv}(t) - \mu_{iv}(t))^2  \nonumber \\
    &-1 +\log (\sigma_{jv}^{2}) - \log (\sigma_{iv}^{2})     \bigg),
\end{align}
where $f^{(i)} =  f(X \: | \: R, \mu_i, \Sigma_i  ) $ is the density given in Eq.~\eqref{eq: diag gmm density}.
The KL-divergence can be made symmetric via the transformation
\begin{equation}
     D^S_{KL}( f^{(i)} \: \| \: f^{(j)} ) =\frac{1}{2} \left( D_{KL}( f^{(i)} \: \| \: f^{(j)} ) + D_{KL}( f^{(j)} \: \| \: f^{(i)} ) \right).
\end{equation}
The underlying idea in our semi-supervised framework is to learn the transformation $W$ for the clusters with only unlabeled points by finding the nearest cluster (in the $D^S_{KL}$-sense)  that contain labeled points. This leads to Algorithm~\ref{alg:algorithm semi-supervised posterior transformation}. 

\begin{algorithm}[t]
\small
\caption{Semi-supervised  posterior transformation}
\label{alg:algorithm semi-supervised posterior transformation}
\begin{algorithmic}[1]
  \Require Posteriors $\{\Pi^{(n)}\}_{n=1}^N$ from mixture models consisting of $q_2$ components, labels $\{y^{(n)}\}_ {n=1}^L$, and hyperparameter $h$. 
  \For{$i = 1, \dots, q_2$, $j = 1, \dots, N_c$}
  \State Compute $W_{ij} = \frac{\sum_{n=1}^N y_j^{(n)} \pi_i^{(n)} }{\sum_{n=1}^N y_j^{(n)}}$.
  \EndFor
  \ForAll{$k$ s.t. $\sum_{j=1}^{N_c} W_{kj} < h$}
  \State Let $ \mathcal{L} = \{l \: s.t. \: \sum_{j=1}^{N_c} W_{lj} \geq h \}$ 
  \State $W_{kj} = W_{lj}$ where $l = \arg\min\limits_{l \in \mathcal{L}} D^S_{KL}( f^{(k)} \: \| \: f^{(l)} ) $.
  \EndFor
  \For{$i = 1, \dots, q_2$, $j = 1, \dots, N_c$}
  \State $W_{ij} = \frac{W_{ij}}{\sum_{j=1}^{N_c} W_{ij} }$.
  \EndFor
  \State Transform training or test posterior via   $\tilde{\Pi} = W^T \Pi   $
  \Ensure Transformed posteriors $\tilde{\Pi}^{(n)} $ 
\end{algorithmic}
\end{algorithm}

\section{Experiments on synthetic and benchmark datasets}
\label{sec: experiments}

The experiments in this paper consists of two parts.
The purpose of the first part was to demonstrate within a controlled environment situations where the proposed TCK$_{IM}$ and ssTCK kernels might prove more useful than the TCK. In the second part (Sec.~\ref{Sec: case study}), we present a case study from a real-world medical application in which we compared to several baseline methods.


In the first part, we considered synthetic and benchmark datasets.
The following experimental setup was considered. 
We performed kernel principal component analysis (KPCA) using time series cluster kernels and let the dimensionality of the embedding be 10.
Thereafter, we trained a kNN-classifier with $k = 1$ on the embedding and evaluated performance in terms of classification accuracy on an independent test set. We let $Q = 30$ and $\mathcal{I}_C = \{N_c,\dots, N_c+20\}$.
An additional hyperparameter $h$ was introduced for ssTCK. We set $h$ to $10^{-1}$ in our experiments.
We also standardized each attribute to zero mean and unit standard deviation.

\subsection{Synthetic example}
To illustrate the effectiveness of the proposed methods, we first considered a controlled experiment in which a synthetic MTS dataset with two classes was sampled from a first-order vector autoregressive model, 
\begin{align}
\begin{pmatrix}
x_1(t) \\
x_2(t)
\end{pmatrix}
=
\begin{pmatrix}
\alpha_1 \\
\alpha_2
\end{pmatrix}
+
\begin{pmatrix}
\rho_1 & 0\\
0 & \rho_2
\end{pmatrix}
\begin{pmatrix}
x_1(t-1) \\
x_2(t-1)
\end{pmatrix}
+
\begin{pmatrix}
\xi_1(t) \\
\xi_2(t)
\end{pmatrix}
\end{align}
To make $x_1(t)$ and $x_2(t)$ correlated with $\mathrm{corr}(x_1(t),x_2(t)) = \rho$, we chose the noise term s.t., 
$
\mathrm{corr}\left(\xi_1(t),\xi_2(t)\right) = \rho \: (1- \rho_1 \rho_2) \: [(1-\rho_1^2)(1-\rho_2^2)]^{-1}.
$
For the first class ($y=1$), we generated 100 two-variate MTS of length 50 for the training and 100 for the test, from the VAR(1)-model with parameters $\rho = \rho_1 = \rho_2 = 0.8$ and $\mathbb{E}[(x_1(t), x_2(t))^T \: | \: y = 1] = (0.5, -0.5)^T$.
Analogously, the MTS of the second class ($y=2$) were generated using parameters $\rho = -0.8$, $\rho_1 = \rho_2 = 0.6$ and $\mathbb{E}[(x_1(t), x_2(t))^T \: | \: y = 2] = (0, 0)^T$.

To simulate MNAR and inject informative missing patterns,  we let $x^{(n)}_i(t)$ have a probability $p^{(n)}$ of being missing, given that $x^{(n)}_i(t)>-1$, $i = 1,2$. We let $p^{(n)} = 0.9 $ if $y^{(n)} = 1$ and $p_{(n)} = 0.8$ otherwise. By doing so, the missing ratio was roughly $63 \%$ in both classes.

\begin{figure*}[t]
    \centering
     \subfigure[TCK]
    {
        \includegraphics[trim = {42mm 85mm 42mm 57mm}, clip, width=0.32\linewidth]{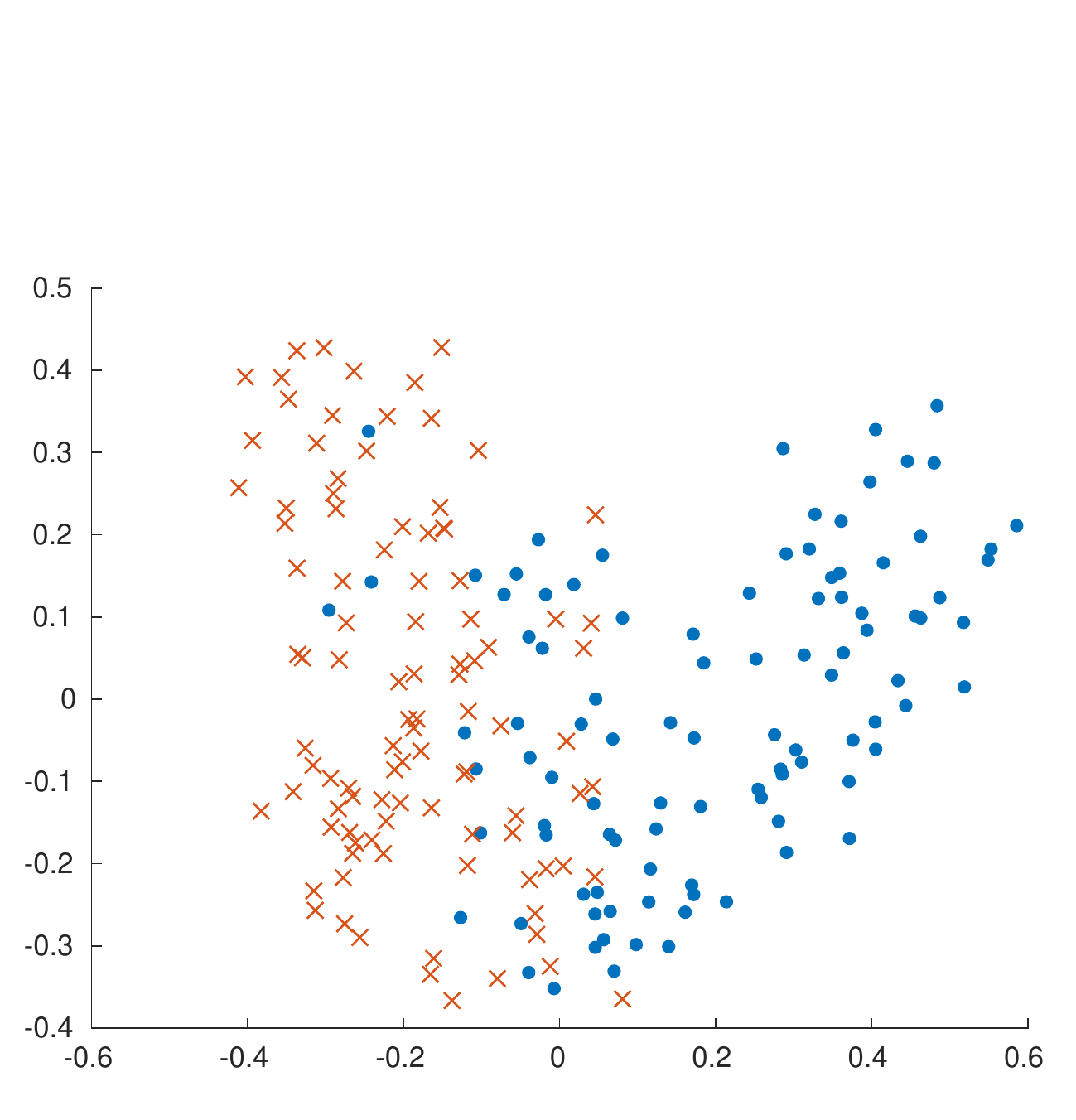}
        \label{fig: var1 u}
    }%
     \subfigure[ssTCK]
    {
        \includegraphics[trim = {42mm 85mm 42mm 57mm}, clip, width=0.32\linewidth]{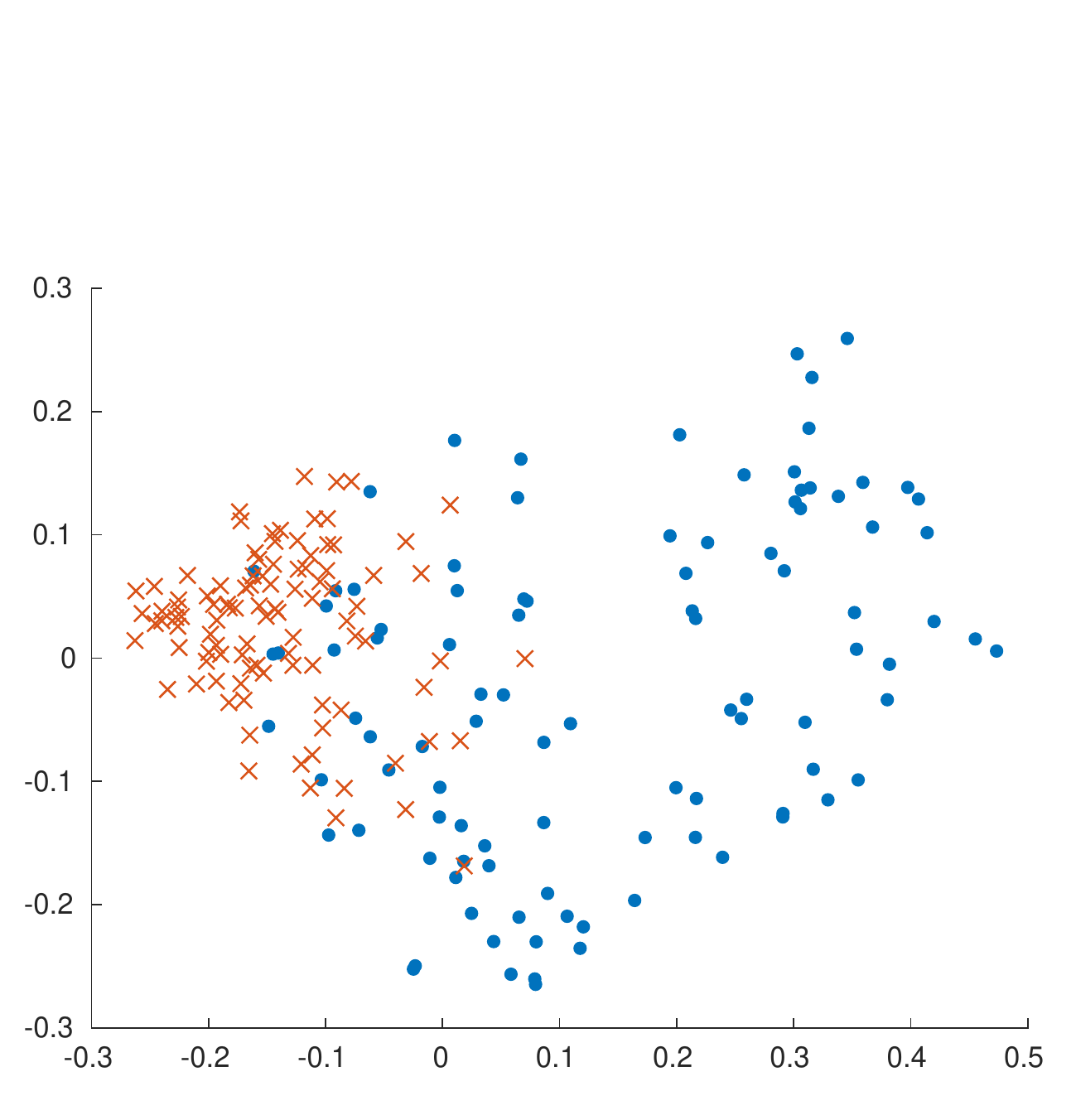}
        \label{fig: var1 ss}
    }%
   \subfigure[sTCK]
    {
        \includegraphics[trim = {42mm 82mm 42mm 95mm}, clip, width=0.32\linewidth]{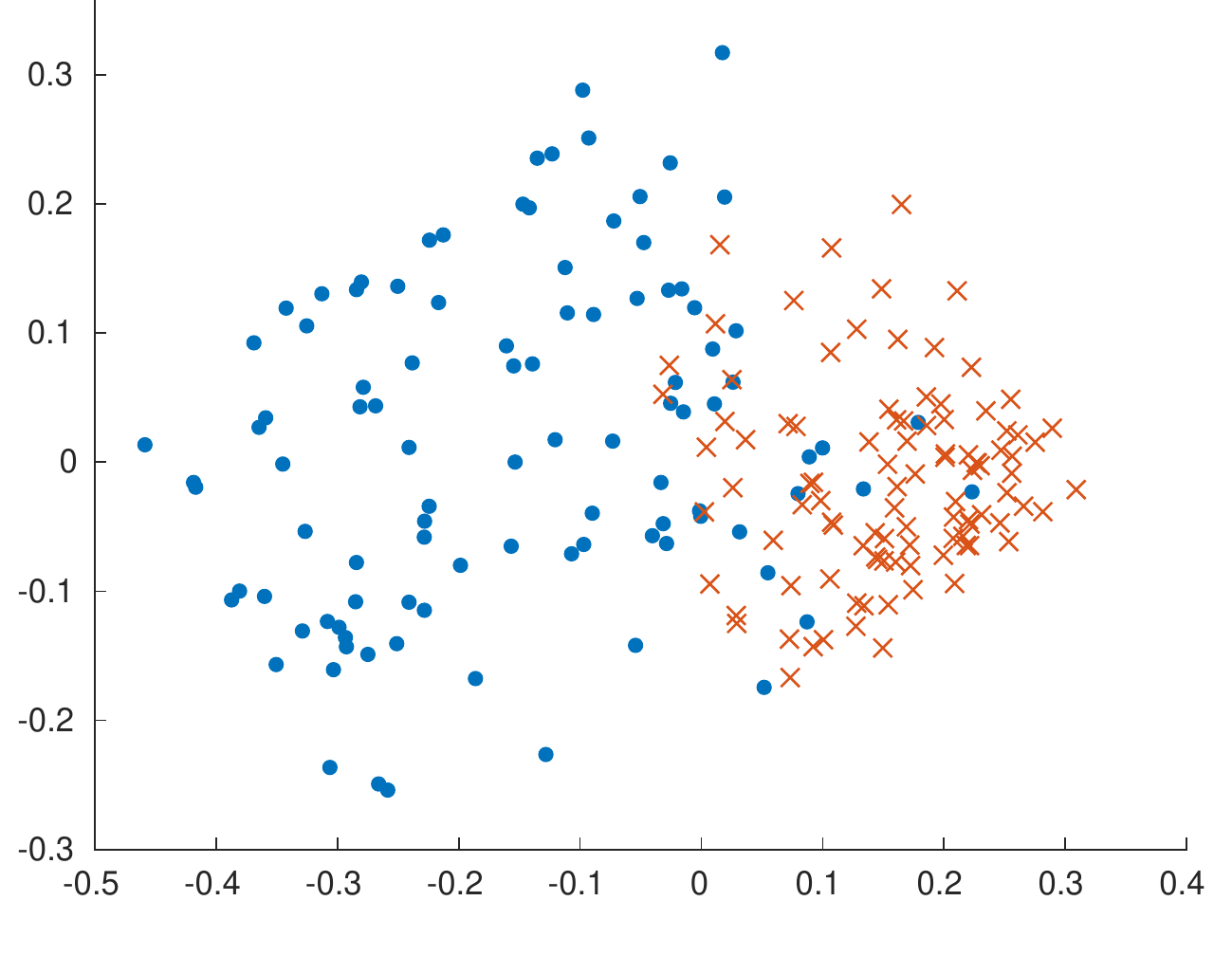}
        \label{fig: var1 s}
    } 
         \subfigure[TCK$_{IM}$]
    {
        \includegraphics[trim = {42mm 85mm 42mm 57mm}, clip, width=0.32\linewidth]{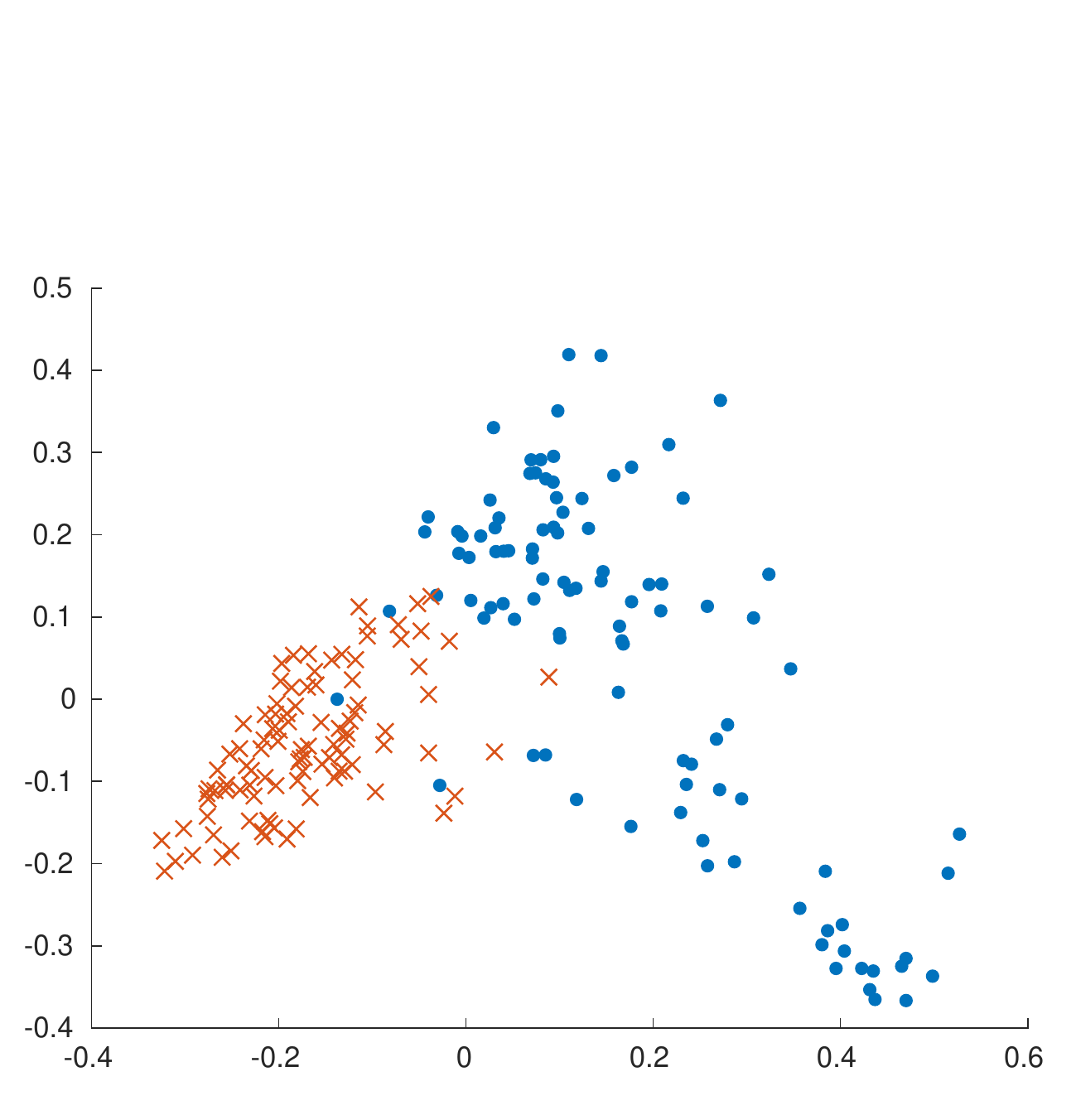}
        \label{fig: var1 im u}
    }%
         \subfigure[ssTCK$_{IM}$]
    {
        \includegraphics[trim = {42mm 85mm 42mm 57mm}, clip, width=0.32\linewidth]{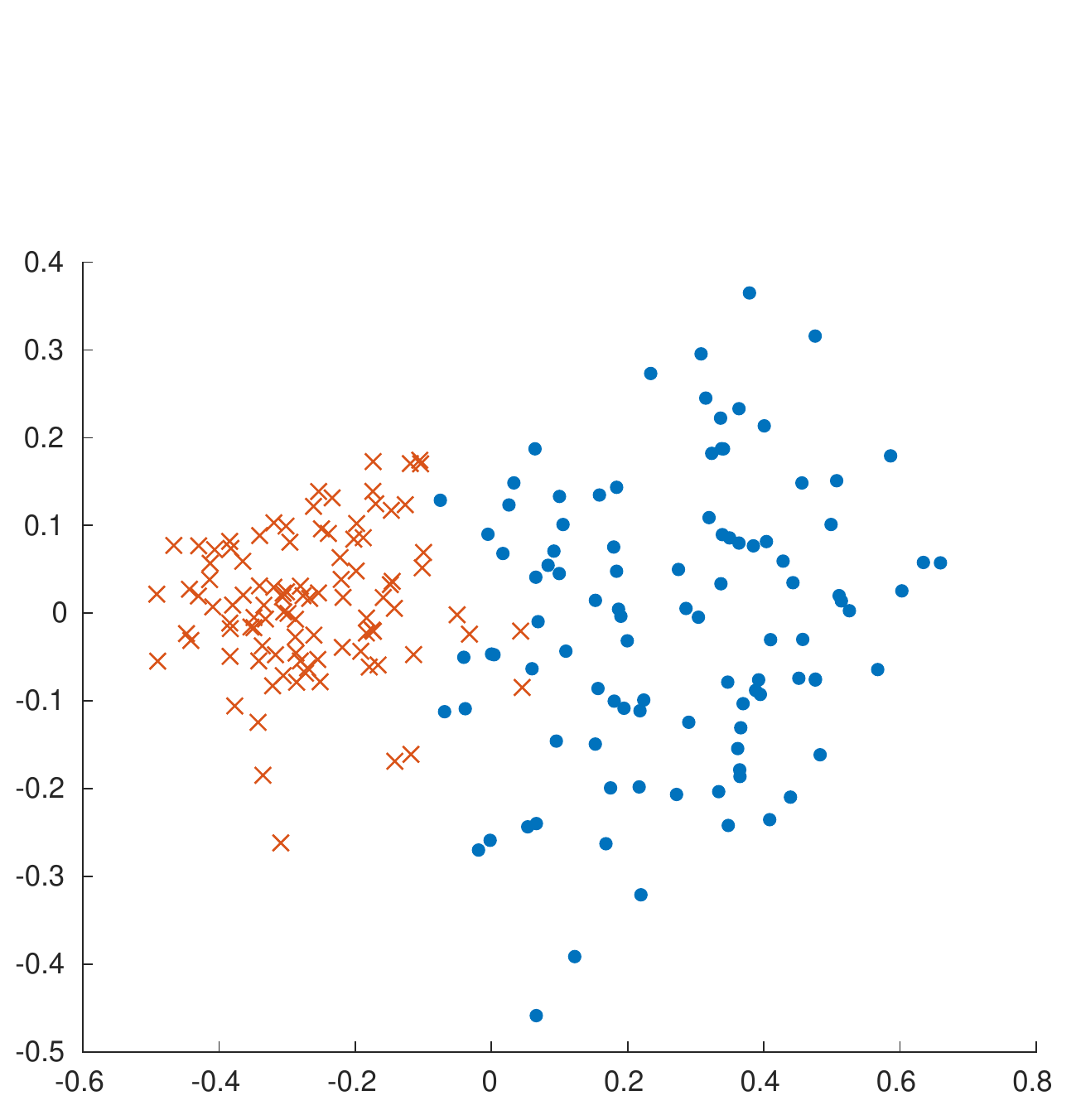}
        \label{fig: var1 im ss}
    }%
         \subfigure[sTCK$_{IM}$]
    {
        \includegraphics[trim = {42mm 85mm 42mm 57mm}, clip, width=0.32\linewidth]{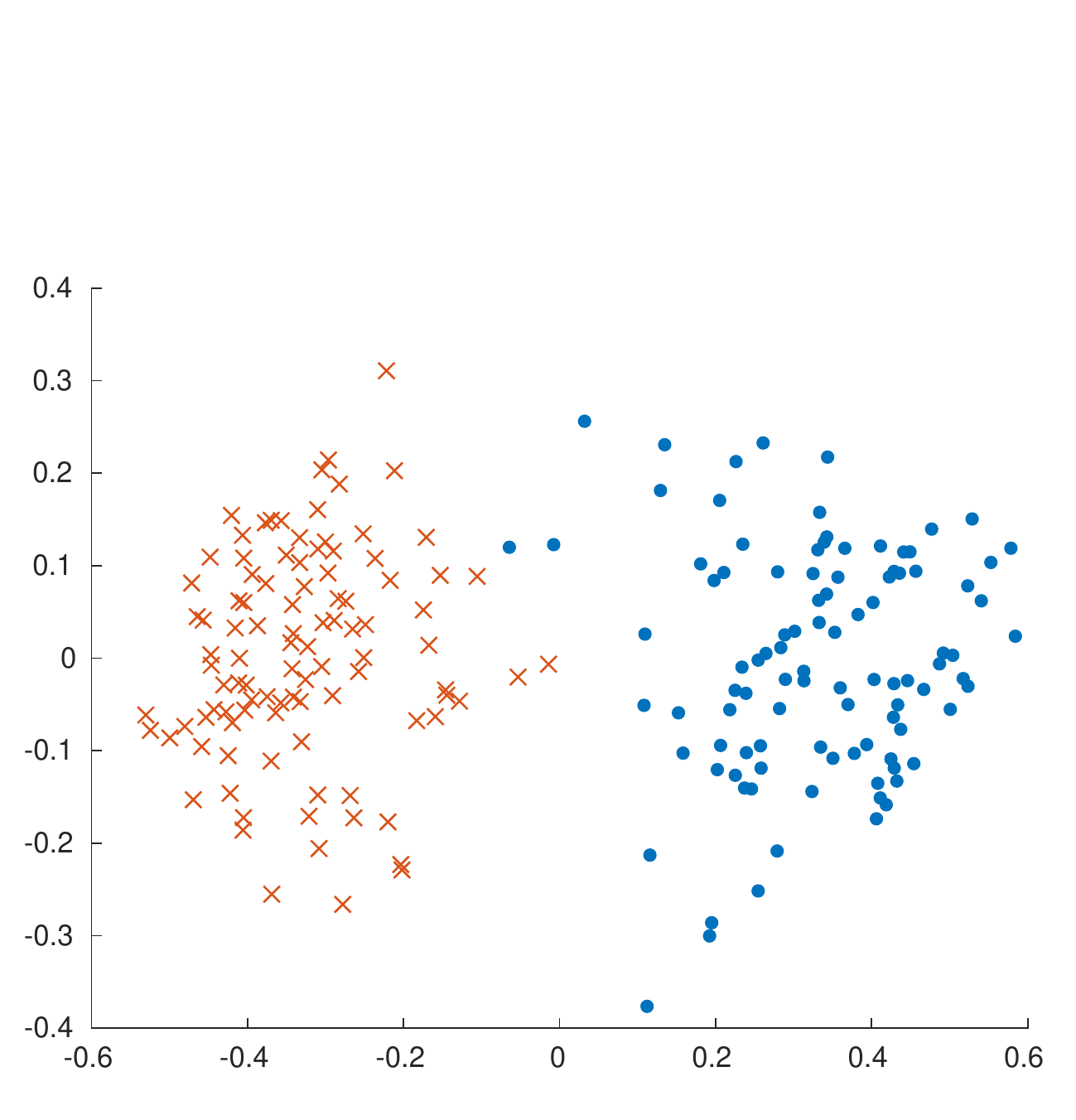}
        \label{fig: var1 im s}
    } 
    \caption{Plot of the two-dimensional KPCA representation of the synthetic data obtained using 6 different time series cluster kernels. The datapoints are color-coded according to their labels.}
    \label{fig: var1}
\end{figure*}

\if{
\begin{table}[t]
    \small
    \centering
     \caption{Accuracy on the synthetic VAR(1) dataset.}
    \label{tab:VAR1}
    \begin{tabular}{lccc}
    \hline
     & Unsupervised & Semi-supervised & Supervised \\
     \hline
   TCK     &  0.826 $\pm$ 0.008   &  0.854 $\pm$  0.011 & 0.867 $\pm$  0.007 \\
   TCK$_{IM}$ & 0.933 $\pm$ 0.007 & 0.967 $\pm$  0.004 & 0.970 $\pm$  0.004 \\
    \hline
    \end{tabular}
\end{table}
}\fi

\begin{table}[t]
    \small
    \centering
     \caption{Accuracy on the synthetic VAR(1) dataset.}
    \label{tab:VAR1}
    \begin{tabular}{lccc}
    \hline
     & Unsupervised & Semi-supervised & Supervised \\
     \hline
   TCK     &  0.826   &  0.854 & 0.867 \\
   TCK$_{IM}$ & 0.933  & 0.967  & 0.970  \\
    \hline
    \end{tabular}
\end{table}

Tab.~\ref{tab:VAR1} shows the accuracy on the test data for the different  kernels. As expected, the TCK gives the lowest accuracy, 0.826. The ssTCK improves the accuracy considerably (0.854), and the supervised version (sTCK) gives further improvement (0.867). However, as we can see, the effect of explicitly modeling the missingness mechanism in the TCK$_{IM}$ is larger. In this case the accuracy increases from 0.826 to 0.933. The two corresponding embeddings are plotted in Fig.~\ref{fig: var1 u} and~\ref{fig: var1 im u}, respectively. In the TCK embedding, there are many points from different classes that overlap with each other, whereas for the TCK$_{IM}$ the number of overlapping points is much lower.

The ssTCK$_{IM}$ improves the accuracy to 0.967 (from 0.933 for TCK$_{IM}$ and 0.854 for ssTCK). The two embeddings obtained using the semi-supervised methods are shown in Fig.~\ref{fig: var1 ss} and~\ref{fig: var1 im ss}.
The supervised version sTCK$_{IM}$ yields a slight improvement in terms of accuracy compared to ssTCK$_{IM}$ (0.970 vs 0.967). Plots of the supervised embeddings are shown in Fig.~\ref{fig: var1 s} and ~\ref{fig: var1 im s}. We can see that for the sTCK$_{IM}$ the classes are clearly separated.

\subsection{Performance of ssTCK on benchmark datasets}
The purpose of the experiments reported in the following paragraph was to evaluate the impact of incorporating incomplete label information in the ssTCK. Towards that end, we considered benchmark datasets and artificially modified the number of labeled MTS in the training sets. 
We applied the proposed ssTCK to four MTS benchmark datasets from the UCR and UCI databases~\cite{UCRArchive, Lichman:2013} and other published work~\cite{ Olszewski}, described in Tab.~\ref{tab: benchmark description}. 
Since some of the datasets contain MTS of varying length, we followed the approach of Wang et al.~\cite{Wang2016237} and transformed all the MTS in the same dataset to the same length, $T$, determined by 
$
T = \left \lceil \frac{T_{max}}{\left \lceil \frac{T_{max}}{25} \right \rceil} \right \rceil,
$
where $T_{max}$ is the length of the longest MTS in the dataset and $ \lceil \: \rceil$ is the ceiling operator. 
The number of labeled MTS was set to $\max \{20, 3 \cdot N_c \} $. ssTCK was compared to ordinary TCK and sTCK (assuming complete label information in the latter case).

\begin{table}[!tb]
\small
\centering
\caption{Description of benchmark time series datasets. Column 2 to 5 show the number of attributes, samples in training and test set, and number of classes, respectively. $T_{min}$ is the length of the shortest MTS in the dataset and $T_{max}$ the longest MTS. $T$ is the length of the MTS after the transformation.}\label{tab: benchmark description}
\begin{tabular}{@{}l@{}c@{\:}@{\:}ccccc@{\:}@{\:}c@{\:}@{\:}r@{}}
\hline
Datasets & Attributes &  Train  & Test  &  $N_c$  & $T_{min}$ & $T_{max}$ & $T$ & Source \\
\hline
uWave & 3 & 200 & 4278 & 8 & 315 & 315 & 25 & UCR \\
Char.Traj. & 3 & 300 & 2558 & 20 & 109 & 205 & 23 & UCI \\
Wafer & 6 & 298 & 896 & 2 & 104 & 198 & 25 & Olsz. \\
Japan.vow. & 12 & 270 & 370 & 9 & 7 & 29 & 15 & UCI \\
\hline
\end{tabular}
\end{table}

\begin{table}[!tb]
\small
\centering
\caption{Classification accuracy for benchmark datasets obtained using TCK, ssTCK and sTCK.}
\label{tab: ss and s benchmark results}
\begin{tabular}{llll}
\hline
 Datasets &  TCK  & ssTCK  & sTCK  \\
\hline
Char. Traj.  & 0.908  & 0.928  &   0.934 \\
 uWave  & 0.867  & 0.881  &   0.894  \\
Wafer    & 0.956  & 0.970  &   0.970  \\
Japanese vowels   & 0.946 & 0.962  &   0.968  \\ 
\hline
\end{tabular}
\end{table}

Tab.~\ref{tab: ss and s benchmark results} shows the performance of ssTCK for the 4 benchmark datasets. As we can see, compared to TCK, the accuracy in general increases using ssTCK. For the Wafer dataset, ssTCK yields the same performance as the supervised kernel. For the three other datasets, the performance of ssTCK is slightly worse than sTCK. These experiments demonstrate that ssTCK is capable of exploiting incomplete label information.

\begin{table*}[!tb]
\small
\centering
\caption{Classification accuracy for benchmark datasets obtained using TCK, ssTCK and sTCK.}
\label{tab: ss and s benchmark results 2}
\begin{tabular}{ll|lll|lll}
\hline
Missing rate & Datasets &  TCK  & ssTCK  & sTCK & GAK &  Linear  & LPS \\
\hline
50\%  &Char. Traj.  & 0.751 &	0.780 &	0.797 & 0.588  & 0.589    & 0.127\\
& uWave & 0.812 &	0.834 &	0.850 &  0.828  &  0.813   & 0.411\\
&Wafer    & 0.956  & 0.970  &   0.972  & 0.792   & 0.791 &  0.823\\
&Japanese vowels   & 0.929 & 0.948  &   0.947 &   0.827 &       0.824 & 0.746 \\ 
\hline
80\%  &Char. Traj.  & 0.282 &	0.310 &	0.331 & 0.194 &      0.192  &  0.062\\
& uWave & 0.589 &	0.592  &	0.603   & 0.441    & 0.464    & 0.234\\
&Wafer    & 0.926 &	0.934 &	0.934   & 0.796  &     0.805 &  0.819 \\
&Japanese vowels   & 0.809 & 0.836  &   0.847  & 0.473   & 0.489 & 0.389 \\ 
\hline
\end{tabular}
\end{table*}

\begin{table*}[!t]
\small
\centering
\caption{Classification accuracy on synthetic benchmark datasets that contain missing data.}
\label{tab: im synthetic results}
\begin{tabular}{l|llll|llll}
\hline
\hline
Correlation &  TCK  & TCK$_{B}$ & TCK$_{0}$ &  TCK$_{IM}$   &  TCK  & TCK$_{B}$ & TCK$_{0}$ &  TCK$_{IM}$ \\
\hline
\hline
 &  \multicolumn{4}{c|}{\textbf{Wafer}} & \multicolumn{4}{c}{\textbf{Japanese vowels}} \\
0.2  & 0.951  & 0.951  & 0.951  & \textbf{0.955}  &  0.938 &  \textbf{0.954} & 0.951 &  0.940   \\
0.4 & \textbf{0.961}  & 0.953  & 0.955  & \textbf{0.961}  &  0.932 &  0.938 & 0.938 &  \textbf{0.941}   \\
0.6 & 0.961  & 0.900  & 0.965  & \textbf{0.996}  &  0.922 &  0.946 & 0.924 &  \textbf{0.962}   \\
0.8  & 0.958  & 0.893  & 0.963  & \textbf{1.000}  &  0.922 &   0.924  &   0.935  &  \textbf{0.968}  \\
\hline
\hline
&  \multicolumn{4}{c|}{\textbf{uWave}} & \multicolumn{4}{c}{\textbf{Character trajectories}} \\
0.2 & 0.763 & 0.457 &  0.755 & \textbf{0.841} & \textbf{0.854} & 0.742 & 0.847 & 0.851 \\
0.4 & 0.807 & 0.587 & 0.813 & \textbf{0.857} & 0.851 & 0.788 & 0.842 & \textbf{0.867}\\
0.6 & 0.831 & 0.674 & 0.837 & \textbf{0.865} & 0.825 & 0.790 & 0.824 & \textbf{0.871} \\
0.8 & 0.834 & 0.699 & 0.844 & \textbf{0.884} & 0.839 & 0.707 & 0.853 & \textbf{0.901} \\
\hline
\hline
\end{tabular}
\end{table*}

Further, we created 8 synthetic datasets by randomly removing 50\% and 80\%, respectively, of the values in each of the 4 benchmark datasets. As we can see from the results presented in Tab.~\ref{tab: ss and s benchmark results 2}, also in presence of missing data the accuracy in general increases using ssTCK, compared to TCK. 

For comparison, in Tab.~\ref{tab: ss and s benchmark results 2} we also added the results obtained using three other kernels; GAK, the linear kernel, and LPS.  GAK and the linear kernel cannot process incomplete MTS and therefore we created complete datasets using mean imputation for these two kernels. LPS\footnote{Matlab implementation: http://www.mustafabaydogan.com/} was run using default hyperparameters
, with the exception that we adjusted the segment length to be sampled from the interval $[6, 0.8T]$ to account for the relatively short MTS in our datasets. In accordance with~\cite{Cuturi}, for GAK\footnote{Matlab implementation: http://www.marcocuturi.net/GA.html}  we set the bandwidth $\sigma$ to 0.1 times the median  distance of all MTS  in the training set scaled by the square root of the median length of all MTS, and the triangular parameter to 0.2 times the median length of all MTS. Distances were measured using the canonical metric induced by the Frobenius norm. In the linear kernel we set the constant $c$ to 0.  As we can see, the performance of these kernels is considerably worse than the time series cluster kernels for 7 out of 8 datasets. For uWave with 50\% missingness, the performance of GAK and the linear kernel is similar to the TCK kernels. 

\if{
\begin{table}[!tb]
\small
\centering
\caption{Classification accuracy for benchmark datasets obtained using TCK, ssTCK and sTCK.}
\label{tab: ss and s benchmark results}
\begin{tabular}{llll}
\hline
Datasets &  TCK  & ssTCK  & sTCK  \\
\hline
Char. Traj.  & 0.908 $\pm$ 0.010 & 0.928 $\pm$ 0.006 &   0.934 $\pm$ 0.007 \\
 uWave & 0.874 $\pm$ 0.006 &	0.892 $\pm$ 0.006 &	0.903 $\pm$ 0.003  \\
Wafer    & 0.956 $\pm$ 0.007 & 0.970 $\pm$ 0.006 &   0.970 $\pm$ 0.004 \\
Japanese vowels   & 0.946 $\pm$ 0.007 & 0.962 $\pm$ 0.006 &   0.968 $\pm$ 0.004 \\ 
\hline
\end{tabular}
\end{table}
}\fi

\if{
\begin{table}[!tb]
\small
\centering
\caption{Classification accuracy for benchmark datasets obtained using TCK, ssTCK and sTCK.}
\label{tab: ss and s benchmark results 2}
\begin{tabular}{lllll}
\hline
Missing rate & Datasets &  TCK  & ssTCK  & sTCK  \\
\hline
0\% &Char. Traj.  & 0.908  & 0.928  &   0.934 \\
 & uWave & 0.874 &	0.892  &	0.903   \\
&Wafer    & 0.956  & 0.970  &   0.970  \\
&Japanese vowels   & 0.946 & 0.962  &   0.968  \\ 
\hline
50\%  &Char. Traj.  & 0.751 &	0.780 &	0.797\\
& uWave & 0.876 &	0.885 &	0.900  \\
&Wafer    & 0.956  & 0.970  &   0.972  \\
&Japanese vowels   & 0.929 & 0.948  &   0.947  \\ 
\hline
80\%  &Char. Traj.  & 0.282 &	0.310 &	0.331\\
& uWave & 0.854 &	0.860  &	0.876   \\
&Wafer    & 0.926 &	0.934 &	0.934  \\
&Japanese vowels   & 0.809 & 0.836  &   0.847  \\ 
\hline
\end{tabular}
\end{table}
}\fi

\if{
\begin{table*}[!tb]
\small
\centering
\caption{Classification accuracy for benchmark datasets obtained using TCK, ssTCK and sTCK.}
\label{tab: ss and s benchmark results}
\begin{tabular}{ll|lll|lllll}
\hline
Miss. rate & Datasets &  TCK  & ssTCK  & sTCK & GAK$_{mean}$ & GAK$_{LOCF}$ & Lin$_{mean}$ & Lin$_{LOCF}$ & LPS \\
\hline
20\% &Char. Traj.  & 0.848  & 0.867  &   0.877 & 0.872 & 0.890 & 0.861  &  0.895 & 0.392\\
 & uWave & 0.854 &	0.879  &	0.897  &   0.891 & 0.896 & 0.885 &   0.896 & 0.696 \\
&Wafer    & 0.968  & 0.973  &   0.978 & 0.849 & 0.859  &0.852  &   0.855 & 0.904 \\
&Japanese vowels   & 0.940 & 0.956  &   0.964  & 0.913 & 0.948 & 0.913&     0.951 & 0.876\\ 
\hline
50\%  &Char. Traj.  & 0.751 &	0.780 &	0.797 & 0.588  &  0.654 & 0.589  & 0.611  & 0.127\\
& uWave & 0.812 &	0.834 &	0.850 & 0.828 &   0.849 &  0.813   & 0.832 & 0.411\\
&Wafer    & 0.956  & 0.970  &   0.972  & 0.792  &  0.839 & 0.791 & 0.842 & 0.823\\
&Japanese vowels   & 0.929 & 0.948  &   0.947 &   0.827 &    0.937  &    0.824 & 0.940 &0.746 \\ 
\hline
80\%  &Char. Traj.  & 0.282 &	0.310 &	0.331 & 0.194 &   0.174 &   0.192  & 0.173& 0.062\\
& uWave & 0.589 &	0.592  &	0.603  & 0.441  &  0.481  & 0.464   & 0.476  & 0.234\\
&Wafer    & 0.926 &	0.934 &	0.934   & 0.796  &     0.818 & 0.805 & 0.815 & 0.819 \\
&Japanese vowels   & 0.809 & 0.836  &   0.847  & 0.473 &   0.786  & 0.489 & 0.773 & 0.389 \\ 
\hline
\end{tabular}
\end{table*}
}\fi

\subsection{Exploiting informative missingness in synthetic benchmark datasets}
\if{
\begin{table*}[!t]
\small
\centering
\caption{Classification accuracy on synthetic benchmark datasets that contain MNAR missing data.}
\label{tab: im synthetic results MNAR}
\begin{tabular}{l|llll|llll}
\hline
Correlation &  TCK  & TCK$_{B}$ & TCK$_{0}$ &  TCK$_{IM}$   &  TCK  & TCK$_{B}$ & TCK$_{0}$ &  TCK$_{IM}$ \\
\hline
&  \multicolumn{4}{c|}{\textbf{uWave}} & \multicolumn{4}{c}{\textbf{Character trajectories}} \\
0.2 & 0.852 & 0.631 &  0.846 & \textbf{0.856} & \textbf{0.854} & 0.670 & 0.847 & 0.848 \\
0.4 & 0.863 & 0.567 & 0.852 & \textbf{0.869} & 0.837 & 0.614 & 0.814 & \textbf{0.870 }\\
0.6 & 0.858 & 0.537 & 0.865 & \textbf{0.870} & 0.856 & 0.678 & 0.845 & \textbf{0.876} \\
0.8 & 0.850 & 0.587 & 0.856 & \textbf{0.874} & 0.829 & 0.542 & 0.850 & \textbf{0.883} \\
\hline
\end{tabular}
\end{table*}
}\fi

\if{
\begin{table*}[!t]
\small
\centering
\caption{Classification accuracy on synthetic benchmark datasets that contain MNAR missing data.}
\label{tab: im synthetic results MNAR}
\begin{tabular}{l|llll|llll}
\hline
Correlation &  TCK  & TCK$_{B}$ & TCK$_{0}$ &  TCK$_{IM}$   &  TCK  & TCK$_{B}$ & TCK$_{0}$ &  TCK$_{IM}$ \\
\hline
&  \multicolumn{4}{c|}{\textbf{uWave}} & \multicolumn{4}{c}{\textbf{Character trajectories}} \\
0.2 & 0.763 & 0.457 &  0.755 & \textbf{0.841} & \textbf{0.854} & 0.742 & 0.847 & 0.851 \\
0.4 & 0.807 & 0.587 & 0.813 & \textbf{0.857} & 0.851 & 0.788 & 0.842 & \textbf{0.867}\\
0.6 & 0.831 & 0.674 & 0.837 & \textbf{0.865} & 0.825 & 0.790 & 0.824 & \textbf{0.871} \\
0.8 & 0.834 & 0.699 & 0.844 & \textbf{0.884} & 0.839 & 0.707 & 0.853 & \textbf{0.901} \\
\hline
\end{tabular}
\end{table*}

}\fi


To evaluate the effect of modeling the missing patterns in TCK$_{IM}$,  we  generated 8 synthetic datasets by manually injecting missing elements into the Wafer and Japanese vowels datasets using the following procedure. For each attribute $v \in \{1,\dots, V\} $, a number $c_v \in \{-1, 1\} $ was randomly sampled with equal probabilities. If $c_v =1$, the attribute $v$ is positively correlated with the labels, otherwise negatively correlated. For each MTS $X^{(n)}$ and attribute, a missing rate $\gamma_{nv}$ was sampled from the uniform distribution  $\mathcal{U}[  0.3 + E \cdot c_v \cdot  (y^{(n)}-1), 0.7 + E \cdot c_v \cdot (y^{(n)}-1)]$. This ensures that the overall missing rate of each dataset is approximately 50\%. $y^{(n)} \in \{1, \dots N_c\}$ is the label of the MTS $X^{(n)}$ and $E$ is a parameter, which we tune for each dataset in such a way that the absolute value of the Pearson correlation between the missing rates for the attributes $\gamma_v$  and the labels $y^{(n)}$ takes the values $\{ 0.2, \: 0.4, \: 0.6, \: 0.8\}$, respectively. The higher the value of the Pearson correlation, the higher is the informative missingness.

Tab.~\ref{tab: im synthetic results} shows the performance of the proposed TCK$_{IM}$ and three baseline models (TCK, TCK$_B$, and TCK$_0$). 
The first baseline is ordinary TCK, which ignores the missingness mechanism. For the Wafer dataset, the performance of this baseline was quite similar across all four settings. For the Japanese vowels dataset, the performance actually decreases as the information in the missing patterns increases. 
In the second baseline, TCK$_B$, we tried to model the missing patterns by concatenating the binary missing indicator MTS $R$ to the MTS $X$ and creating a new MTS with $2V$ attributes. Then, we trained ordinary TCK on this representation. For the Wafer dataset, the performance decreases considerably as the  informative missingness increases. For the Japanese vowels, this baseline yields the best performance when the correlation is $20\%$. However,  the performance actually decreases as the informative missingness increases.
Hence, informative missingness is not captured with this baseline. 
In the last baseline, TCK$_0$, we investigated if it is possible to capture informative missingness by imputing zeros for the missing values and then training the TCK on the imputed data.  This baseline yields similar performance across all 4 settings for the Wafer dataset, and for Japanese vowels,  TCK$_0$ has a similar behaviour as  TCK$_B$, i.e. it does not capture informative missing patterns.
The proposed TCK$_{IM}$ achieves the best accuracy for 7 out of 8 settings and has the expected behaviour, namely that the accuracy increases as the correlation between missing values and class labels increases. The performance is similar to TCK when the amount of information in the missing patterns is low, whereas TCK is clearly outperformed when the informative missingness is high. This demonstrates that TCK$_{IM}$ effectively utilizes informative missing patterns.

\if{
To also test if TCK$_{IM}$ is capable of exploiting other types of informative missingness, we generated 8 synthetic datasets from uWave  and Character trajectories using the following approach.
Both of these datasets consists of 3 attributes. We randomly assigned the three numbers $\{-1, 0, 1\}$ to $c_v$. For the attribute with $c_v = -1$, we let it be negatively correlated with the labels by sampling the missing rate $\gamma_{nv}$  from  
$\mathcal{U}[ \max \{ 0.8 - E \cdot  (y^{(n)}-1), 0\}, \min \{1.2 - E  \cdot (y^{(n)}-1), 1\}]$. For the attribute with $c_v = 1$, we let it be positively correlated with the labels by sampling the missing rate $\gamma_{nv}$  from  
$\mathcal{U}[ \max \{ 0.2 + E \cdot  (y^{(n)}-1), 0\}, \min \{0.6 + E  \cdot (y^{(n)}-1), 1\}]$. For $c_v = 0$, we let the attribute be uncorrelated with the labels by sampling the missing rate $\gamma_{nv}$  from  
$\mathcal{U}[ 0.3 ,0.7]$. We let each element with $x^{(n)}_v(t) > \mu_v$  have a probability $\gamma_{nv}$ of being missing, where $\mu_v$ is the mean of attribute $v$ computed over the complete dataset.
We tuned the parameter $E$ such that the mean Pearson correlation between $\gamma_{-1}$ and the labels, and $\gamma_{1}$ and the labels,  took the values $\{ 0.2, \: 0.4, \: 0.6, \: 0.8\}$. By doing so, the overall missing rate was approximately 30\% for uWave and 37\% for the Characters.

Tab.~\ref{tab: im synthetic results MNAR} shows the performance on the 8 synthetic datasets created from uWave and Char. traj. 
One  thing to notice here is the poor performance of TCK$_B$. This demonstrates the importance of using the mixed mode mixtures to model the two modalities in $U = (X, R)$. To naively apply TCK based on the GMMs to the concatenated MTS do not provide the desired performance.
Further, we see that  TCK$_{IM}$ achieves the best accuracy for 7 out of 8 settings and that also for these datasets the accuracy is similar to TCK for low correlation but increases as the missingness information increases.
}\fi

To also test if TCK$_{IM}$ is capable of exploiting other types of informative missingness, we generated 8 synthetic datasets from uWave  and Character trajectories using the following approach.
Both of these datasets consists of 3 attributes.
For each attribute $v \in \{1,\dots, V\} $, a number $c_v \in \{-1, 1\} $ was randomly sampled with equal probabilities. For the attribute(s) with $c_v = -1$, we let it be negatively correlated with the labels by sampling the missing rate $\gamma_{nv}$  from  
$\mathcal{U}[ 0.7 - E \cdot  (y^{(n)}-1), 1 - E  \cdot (y^{(n)}-1)]$. For the attribute with $c_v = 1$, we let it be positively correlated with the labels by sampling the missing rate $\gamma_{nv}$  from  
$\mathcal{U}[ 0.3 + E \cdot  (y^{(n)}-1), 0.6 + E  \cdot (y^{(n)}-1)]$.  We let each element with $x^{(n)}_v(t) > \mu_v$  have a probability $\gamma_{nv}$ of being missing, where $\mu_v$ is the mean of attribute $v$ computed over the complete dataset. The fact that the probability of being missing depends on the missing values means that, within each class, the missingness mechanism is MNAR.
We tuned the parameter $E$ such that the mean absolute value of the Pearson correlation between $\gamma_{v}$ and the labels took the values $\{ 0.2, \: 0.4, \: 0.6, \: 0.8\}$. By doing so, the overall missing rate was approximately 32\% for uWave and 45\% for the Characters. However, we note that in this case the overall missing rate varies slightly as a function of the Pearson correlation.

Tab.~\ref{tab: im synthetic results} shows the performance on the 8 synthetic datasets created from uWave and Char. traj. 
One  thing to notice here is the poor performance of TCK$_B$. This demonstrates the importance of using the mixed mode mixtures to model the two modalities in $U = (X, R)$. To naively apply TCK based on the GMMs to the concatenated MTS do not provide the desired performance.
Further, we see that  TCK$_{IM}$ achieves the best accuracy for 7 out of 8 settings and  the accuracy increases as the correlation increases.  
For the Characters, the performance of TCK$_{IM}$ is similar to TCK for low correlation but increases as the missingness information increases, whereas the performance of TCK actually decreases. One possible explanation is that for this dataset, two of the variables were positively correlated with the labels and therefore the missing rate increases with increasing correlation. 
Regarding the results for uWave, it is a bit surprising that the largest difference in performance between TCK and TCK$_{IM}$ occurs when the correlation is lowest. There might be several reasons to this: a peculiarity of the dataset and/or  that the MNAR missingness created missing patterns that negatively affect TCK.

\section{Case study: Detecting infections among patients undergoing colon rectal cancer surgery} \label{Sec: case study}
In this case study, the focus was to detect
Surgical Site Infection (SSI), which is one of the most common types of nosocomial infections~\cite{lewis2013} and  represents up to 30\% of hospital-acquired infections~\cite{magill2012prevalence, de2009surgical}.  The importance of the topic of SSI prediction is reflected in several recent initiatives. 
For instance, the current study is part of a larger research effort by the current team, on SSI prediction and detection of postoperative adverse events related to gastrointestinal surgery within the context of improving the \emph{quality of surgery}~\cite{MIKALSEN2017105,soguero2016support,soguero2016predicting, 8333430, soguero2015data, jensen2017analysis}.  Clearly, the reason for this massive interest is that a reduction in the number of postoperative complications such as SSI will be of great benefit both for the patients and for the society.

Many studies have shown that laboratory tests, and blood tests in particular, are especially important predictors for SSI, both pre- and post-operatively~\cite{silvestre2014, soguero2015data,medina2016,angiolini2016, 8333430,liu2017risk, MUJAGIC2018651, goulart2018early, hu2017strategies,gans2015diagnostic,Sanger2016259}.
Therefore, blood tests provided the basis also for this case study.

\subsection{Data collection} \label{Database}

Ethics approval for the parent study was obtained from the Data Inspectorate and the Ethics Committee at the University Hospital of North Norway (UNN)~\cite{jensen2017analysis}. In~\cite{jensen2017analysis}, a cohort consisting of 7741 patients was identified by extracting the electronic health records for all patients that underwent a gastrointestinal surgical procedure at UNN in the years 2004–2012. 
In this case study, we were particularly interested in detecting SSI, which is an infection particularly associated with colorectal cancer surgery~\cite{lawson2013reliability}. 
Therefore,  patients who did not undergo this type of surgery were excluded, reducing the size of the cohort to 1137 patients. 


In collaboration with a clinician (author A. R.), we extracted data for 11 of the most common blood tests from the patient's EHRs. The value of a patient's blood test, e.g. his or hers hemoglobin level, can be considered as a continuous variable over time. However, blood tests are usually measured on a daily basis, and therefore, for the purpose of the current analysis, we discretized time and let each time interval be one day. 
Hence, the blood samples could naturally be represented as MTS and needed no further feature preprocessing in our framework.

\begin{table}[tb]
\small
\centering
\caption{List of extracted blood tests and their corresponding missing rates.} 
\label{tab: blood names}
\begin{tabular}{clc}
\hline
\textbf{Attribute nr.} & \textbf{Blood test} & \textbf{Missing rate} \\
\hline
1 & Hemoglobin  & 0.646  \\
2 & Leukocytes  &  0.727 \\
3 & C-Reactive Protein & 0.691   \\
4 & Potassium & 0.709 \\
5 & Sodium & 0.712  \\
6 & Creatinine & 0.867 \\
7 & Thrombocytes  & 0.921  \\
8 &Albumin & 0.790  \\
9 &Carbamide & 0.940\\
10 &Glucose & 0.921 \\
11 & Amylase & 0.952 \\ 
\hline
\end{tabular}
\end{table}

\begin{figure*}[t]
    \centering
    \includegraphics[trim = 0mm 0mm 0mm 0mm, clip, width=.9\linewidth]{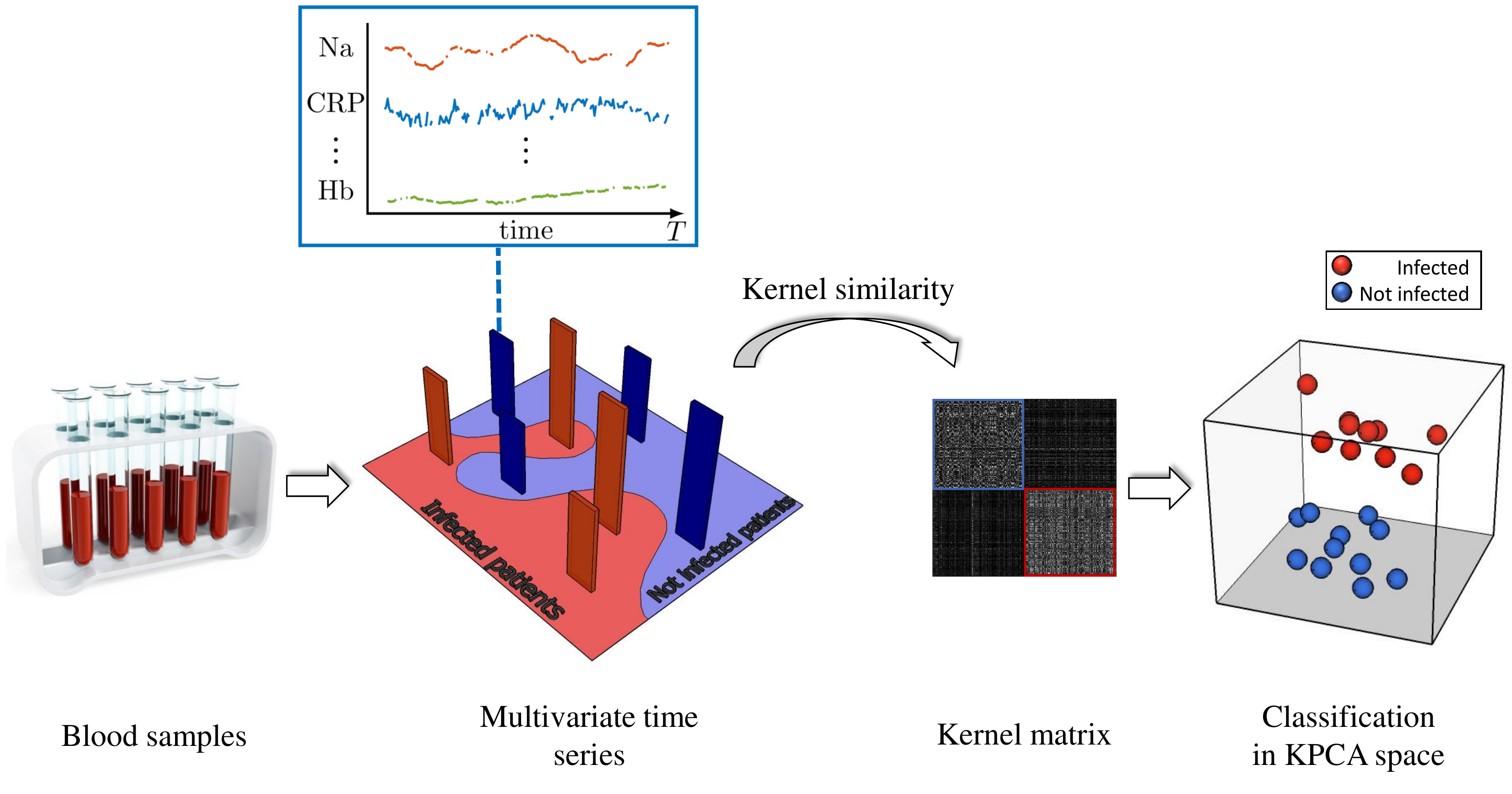}
         \caption{\textit{Overview of the approach taken to detect postoperative SSI from MTS blood samples.}}
    \label{fig:Scheme}
\end{figure*}

All blood tests were not available every day for each patient, which means that the dataset contained missing data, and we expected the missing patterns to be informative since whether a test is performed depends on whether the doctor thinks it is needed. 
We focused on detection of SSI within 10 days after surgery and therefore the length of the time series is 10.
Patients with no recorded lab tests during the period from postoperative day 1 until day 10 were removed from the cohort, which lead to a final cohort consisting of 858 patients. The average proportion of missing data in the cohort was $80.7 \%$.
Tab.~\ref{tab: blood names} shows a list of the blood tests we considered in this study and their corresponding missing rate.

\if{
SSI can be divided into different types depending on the anatomical location of the infection~\cite{ko2015american}.
\emph{Superficial} infections can be treated with local surgical debridement and antibiotics.
On the other hand, \emph{deep} infections are more complex and require lapratomies and/or percutaneous drainage and intravenous antibiotics.
Guided by input from clinicians, the International Classification of Diseases (ICD10) or NOMESCO Classification of Surgical Procedures (NCSP) codes related to severe postoperative complications, both superficial and deep infections, were considered to identify the patients in the cohort that developed postoperative superficial and deep SSI, respectively. Patients that did not have these codes and did not have the word ``infection'' in any of their postoperative text documents were considered as controls. 
For the purpose of this work, and similarly to many earlier studies~\cite{soguero2015data,Sanger2016259,KE201722, LIMON2014127,Gibbons20111,horan_gaynes_martone_jarvis_emori_1992,BERGER2013974}, we did not distinguish between deep and superficial SSI, and therefore we defined the SSI group (cases) as those patients that had any of these two types of SSI.
This lead to a dataset with 227 infected patients (cases) and 631 non-infected ones (control). 
}\fi

Guided by input from clinicians, the International Classification of Diseases (ICD10) or NOMESCO Classification of Surgical Procedures (NCSP) codes related to severe postoperative complications were considered to identify the patients in the cohort that developed postoperative SSI. Patients that did not have these codes and did not have the word ``infection'' in any of their postoperative text documents were considered as controls. This lead to a dataset with 227 infected patients (cases) and 631 non-infected patients (control).

\subsection{Experimental setup}
The objective of this case study was to evaluate how the proposed MTS kernels perform in a real-world application from medicine. We would like to emphasize that the proposed kernels are mainly designed for situations when there are no, or only a few,  ground-truth labels available. However, in order to evaluate the quality of these kernels, we adopted a supervised scheme.  
Hence, we followed the scheme presented in Fig.~\ref{fig:Scheme}, i.e. we computed the kernel from the MTS representations of the blood tests and performed KPCA, followed by kNN classification in the KPCA space.
We set the dimensionality of the KPCA-representation to 10 in all experiments. The number of neighbors $k$ was set using 5-fold cross validation.

Four baseline kernels were considered, namely  TCK, LPS, GAK and the linear kernel. GAK and the linear kernel cannot work on incomplete datasets, and therefore, we created 2 complete datasets using mean and LOCF imputation.
In order to investigate if it is possible to better exploit the information from the missing patterns for the LPS, GAK and linear kernels, we also created baselines by concatenating the binary indicator MTS $R^{(n)}$ to the MTS $X^{(n)}$.

We performed 5-fold cross validation and reported results in terms of F1-score, sensitivity, specificity and accuracy.
Sensitivity is the fraction of actual positives (has SSI) correctly classified as positive, whereas specificity is the fraction of actual negatives that are correctly classified as negative.
F1-score is the harmonic mean of precision and sensitivity, where precision is the fraction of actual positives among all those that are classified  as positive cases.

\begin{table*}[t]
    \centering
        \caption{Performance (mean $\pm$ se) on the SSI dataset.}
        \small
    \begin{tabular}{llcccc}
    \hline
   & Kernel & F1-score & Sensitivity & Specificity & Accuracy \\
     \hline
Ignore &   TCK  & 0.726 $\pm$ 0.045 & 0.678 $\pm$ 0.035 & 0.930 $\pm$ 0.024 & 0.863  $\pm$ 0.023   \\
missingness &   LPS  & 0.746 $\pm$ 0.035 & 0.696 $\pm$ 0.056 & 0.939 $\pm$ 0.019 & 0.875 $\pm$ 0.016  \\
   \hline
Impute &    GAK$_{LOCF}$ & 0.570 $\pm$ 0.045 & 0.484 $\pm$ 0.059 & 0.924 $\pm$ 0.022 & 0.808 $\pm$ 0.017 \\
 &   GAK$_{mean}$ & 0.629 $\pm$ 0.046 & 0.502 $\pm$ 0.059 & 0.966 $\pm$ 0.023 & 0.843 $\pm$0.016 \\
 &   Linear$_{LOCF}$ & 0.557 $\pm$ 0.058 & 0.480 $\pm$ 0.073 & 0.914 $\pm$ 0.017 & 0.800 $\pm$ 0.018  \\
  &  Linear$_{mean}$ & 0.599 $\pm$ 0.030 & 0.489 $\pm$ 0.041 & 0.948 $\pm$ 0.043 & 0.826 $\pm$ 0.024  \\
    \hline
Informative  &   LPS$_{IM}$  & 0.720 $\pm$ 0.062 & 0.661 $\pm$ 0.069 & 0.937 $\pm$ 0.036 & 0.863 $\pm$ 0.032  \\
&    GAK$_{IM + LOCF}$ & 0.669 $\pm$ 0.015 & 0.586 $\pm$ 0.024 & 0.940 $\pm$ 0.021 & 0.846 $\pm$ 0.011 \\
 &   GAK$_{IM + mean}$ & 0.696 $\pm$ 0.030 & 0.617 $\pm$ 0.033 & 0.945 $\pm$ 0.022 & 0.856 $\pm$0.011 \\
 &   Linear$_{IM + LOCF}$ & 0.628 $\pm$ 0.016 & 0.529 $\pm$ 0.030 & 0.945 $\pm$ 0.011 & 0.834 $\pm$ 0.005  \\
  &  Linear$_{IM + mean}$ & 0.668 $\pm$ 0.037 & 0.568 $\pm$ 0.033 & \textbf{0.951 $\pm$ 0.030} & 0.850 $\pm$ 0.021  \\
  &  TCK$_{IM}$  & \textbf{0.802 $\pm$ 0.016} & \textbf{0.806 $\pm$ 0.027} & 0.927 $\pm$ 0.017 & \textbf{0.895 $\pm$ 0.010}  \\
    \hline
    \end{tabular}
    \label{tab:SSI results}
\end{table*}

\if{
\begin{table*}[t]
    \centering
    \small
    \begin{tabular}{llcccc}
    \hline
   & Kernel & F1-score & Sensitivity & Specificity & Accuracy \\
     \hline
Ignore &   TCK  & 0.722 $\pm$ 0.045 & 0.678 $\pm$ 0.035 & 0.927 $\pm$ 0.024 & 0.862  $\pm$ 0.023   \\
missingness &   LPS  & 0.721 $\pm$ 0.035 & 0.678 $\pm$ 0.056 & 0.927 $\pm$ 0.019 & 0.861 $\pm$ 0.016  \\
   \hline
Impute &    GAK$_{LOCF}$ & 0.600 $\pm$ 0.053 & 0.507 $\pm$ 0.066 & 0.936 $\pm$ 0.018 & 0.822 $\pm$ 0.017 \\
 &   GAK$_{mean}$ & 0.575 $\pm$ 0.083 & 0.458 $\pm$ 0.087 & 0.954 $\pm$ 0.010 & 0.823 $\pm$0.027 \\
 &   Linear$_{LOCF}$ & 0.540 $\pm$ 0.074 & 0.440 $\pm$ 0.074 & 0.933 $\pm$ 0.015 & 0.803 $\pm$ 0.027  \\
  &  Linear$_{mean}$ & 0.554 $\pm$ 0.082 & 0.432 $\pm$ 0.088 & \textbf{0.957 $\pm$ 0.010} & 0.818 $\pm$ 0.025  \\
    \hline
Informative  &  TCK$_{IM}$  & \textbf{0.794 $\pm$ 0.022} & \textbf{0.788 $\pm$ 0.035} & 0.929 $\pm$ 0.009 & \textbf{0.892 $\pm$ 0.013}  \\
    \hline
    \end{tabular}
    \caption{Performance (mean $\pm$ se) on the SSI dataset.}
    \label{tab:SSI results}
\end{table*}
}\fi

\begin{figure*}[!t]
    \centering
        \subfigure[LPS]
    {
        \includegraphics[trim = {42mm 93mm 42mm 57mm}, clip, width=0.32\linewidth]{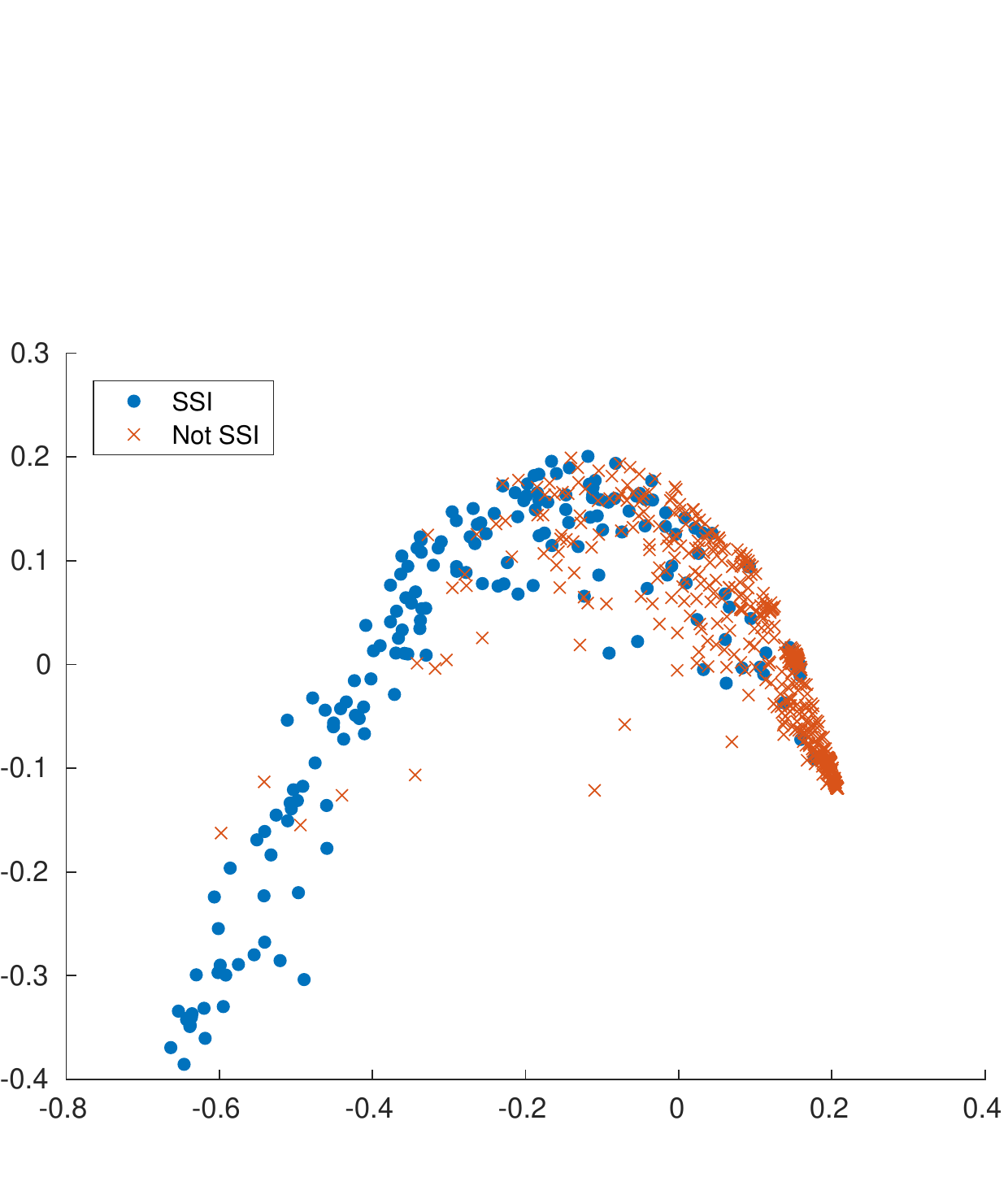}
    }%
    \subfigure[GAK$_{mean}$]
    {
        \includegraphics[trim = {42mm 93mm 42mm 57mm}, clip, width=0.32\linewidth]{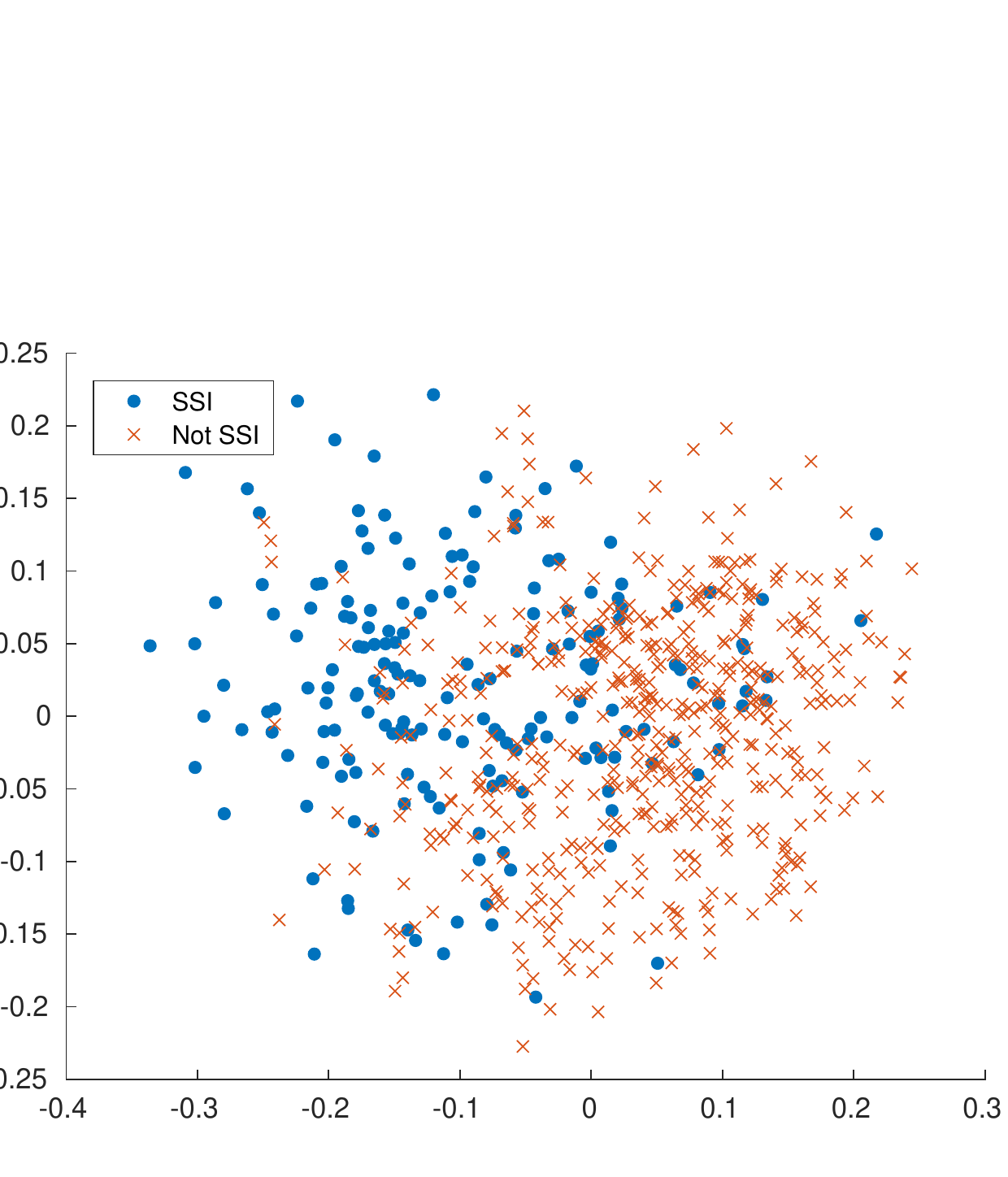}
    }%
    \subfigure[Linear$_{mean}$]
    {
        \includegraphics[trim = {42mm 93mm 42mm 57mm}, clip, width=0.32\linewidth]{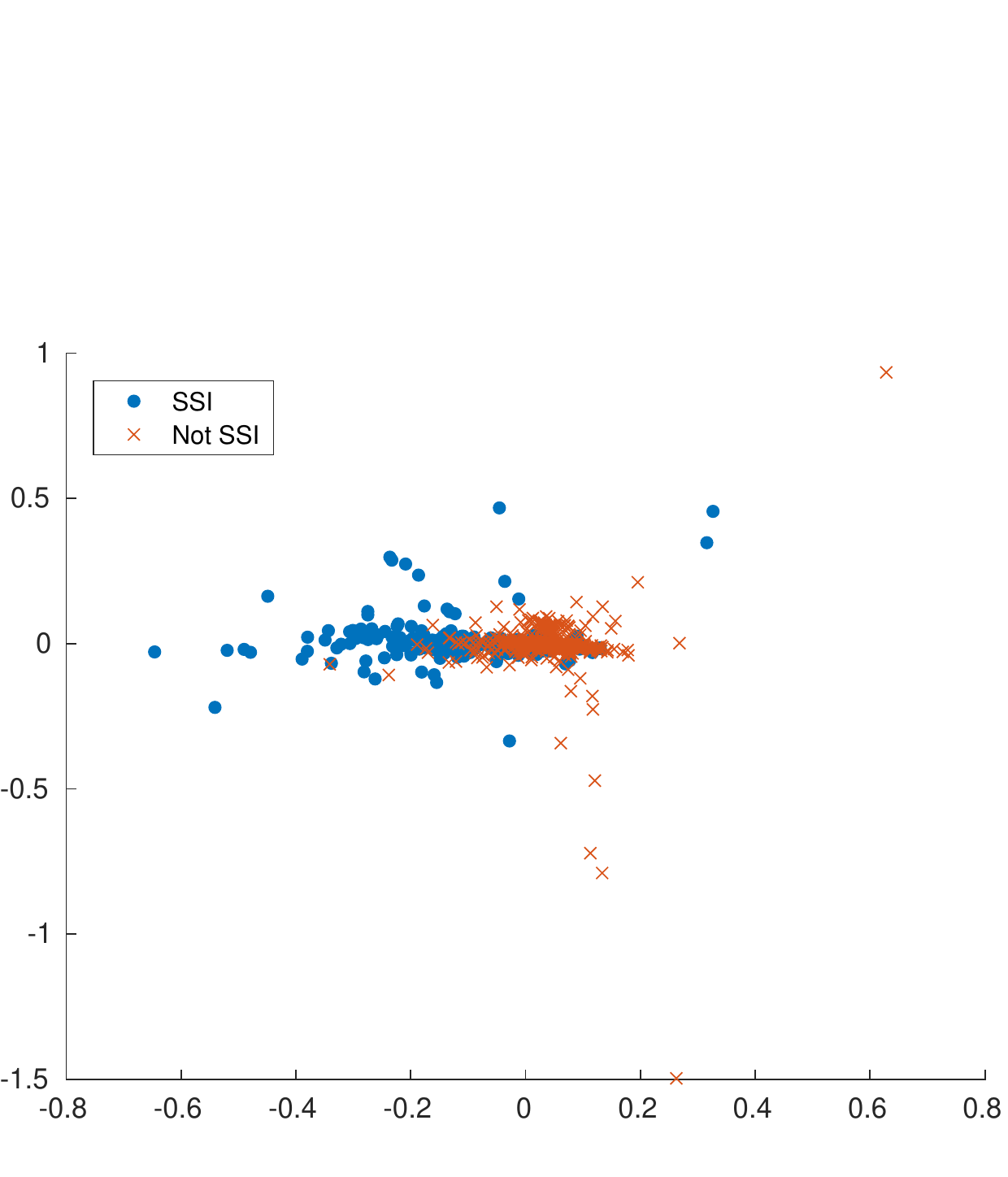}
    }  
        \subfigure[TCK]
    {
        \includegraphics[trim = {42mm 93mm 42mm 57mm}, clip, width=0.32\linewidth]{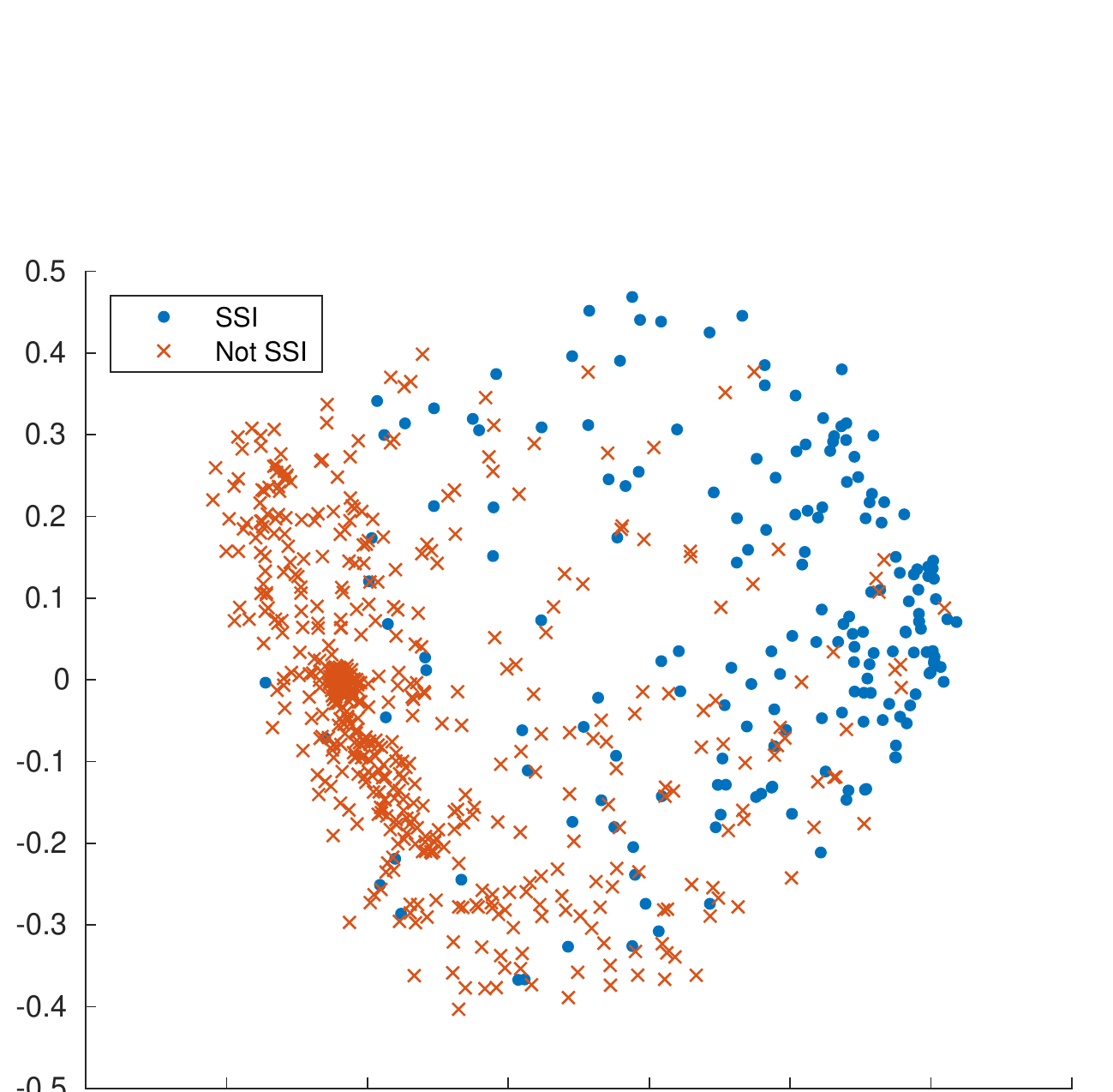}
    }%
    \subfigure[TCK$_{IM}$]
    {
        \includegraphics[trim = {42mm 93mm 42mm 57mm}, clip, width=0.32\linewidth]{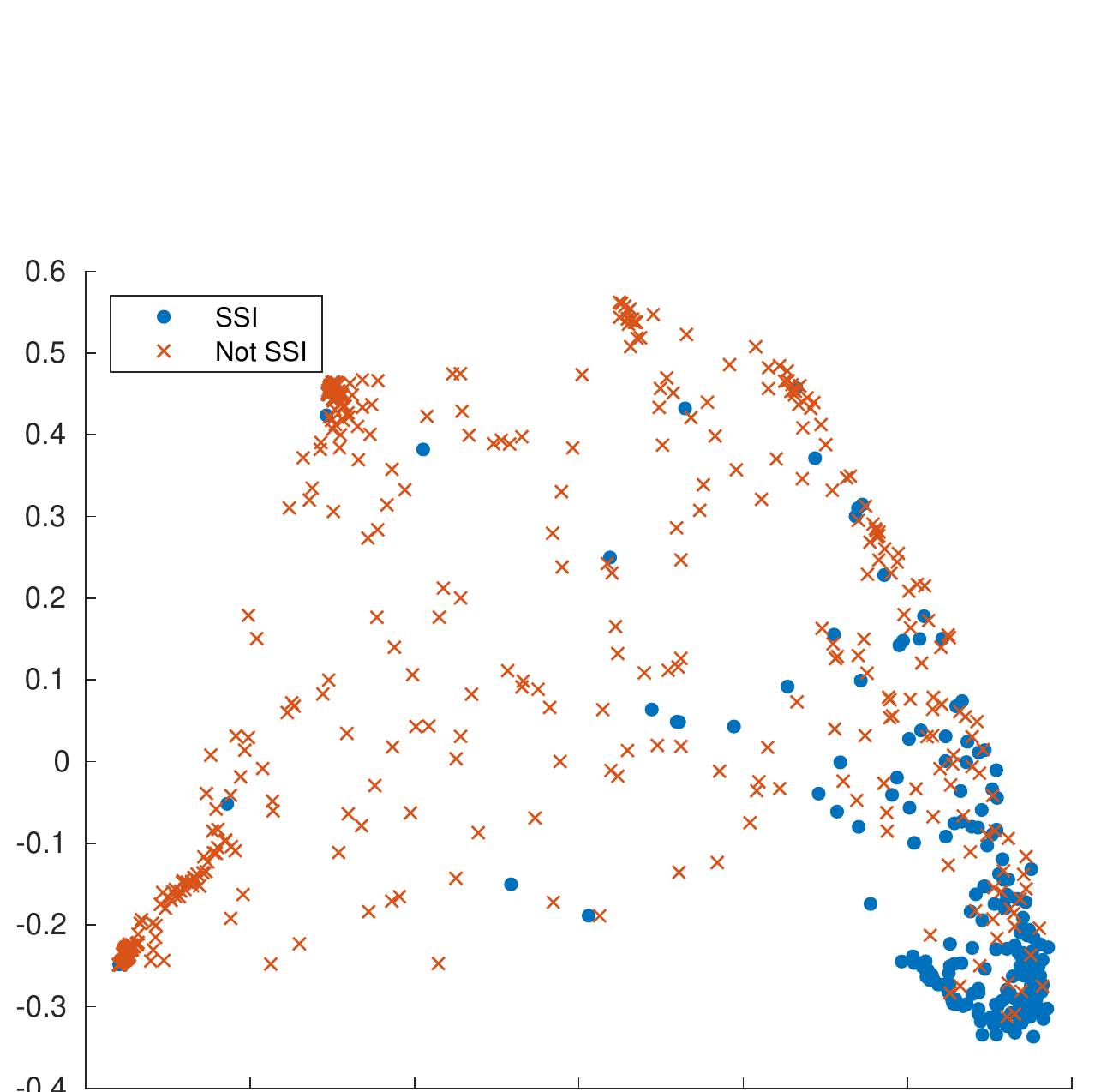}
    }  
    \caption{Plot of the two-dimensional KPCA representation of the colon rectal cancer surgery patients obtained using 5 kernels.}
    \label{fig: SSI plots}
\end{figure*}
\subsection{Results}
Tab.~\ref{tab:SSI results} shows the performance in terms of 4 evaluation metrics for 11 baseline kernels as well as the proposed TCK$_{IM}$ kernel on the task of detecting patients suffering from SSI. 
We see that the kernels that rely on imputation performs much worse than other kernels in terms of F1-score, sensitivity and accuracy. These methods do, however, achieve a high specificity. However, any classifier can achieve a specificity of 1 simply by classifying all cases as negative, but this of course leads to lower F1-score and sensitivity.
The main reasons why these methods do not perform better are probably that the imputation  methods introduce strong biases into the data and that the missingness mechanism is ignored. 
The TCK and LPS kernels perform quite similarly across all 4 evaluation metrics (LPS slightly better). The F1-score, sensitivity and accuracy achieved for these methods are considerably higher than the corresponding scores for the GAK and linear kernel.
One of the reasons why these methods perform better than the imputation methods is that ignoring the missingness leads to lower bias than replacing missing values with biased estimates.
The performance of the linear kernel and GAK improves a bit by accounting for informative missingness, whereas the performance of LPS decreases.
TCK$_{IM}$ performs  similarly to the baselines in terms of specificity, but considerably better in terms of F1-score, sensitivity and accuracy.
This demonstrates that the missing patterns in the blood test time series are informative and the TCK$_{IM}$ is capable of exploiting this information to improve performance on the task of detecting patients with infections.

Fig.~\ref{fig: SSI plots} shows KPCA embeddings corresponding to the two largest eigenvalues obtained using 5 different kernels. While the representations obtained using GAK and the linear kernel are noisy and to a large degree mix the infected and non-infected patients, the two classes (SSI and non-SSI) are more separated in the representations obtained using   TCK and LPS. The TCK$_{IM}$ is even better at forcing the SSI patients to stay in the same region or cluster while it at the same time spreads out the patients without infection, revealing the diversity among these patients.

\if{
\begin{table}[]
    \centering
    \caption{Performance (mean$\pm$se) on SSI dataset for different MTS kernels. }
    \label{tab:VAR1}
    \begin{tabular}{clccc}
    \hline
    &  & Unsupervised & Semi-supervised & Supervised \\
     \hline
   acc   & TCK  & 0.818 $\pm$ 0.007 & 0.791 $\pm$ 0.017  & 0.781 $\pm$ 0.014  \\
   & TCK$_{MNAR}$   & 0.834 $\pm$ 0.007  & 0.843 $\pm$ 0.007 & 0.855 $\pm$ 0.007\\
   \hline
   f1 &  TCK & 0.766 $\pm$ 0.008  & 0.739 $\pm$ 0.022  & 0.731 $\pm$ 0.023 \\
   &   TCK$_{MNAR}$  & 0.809 $\pm$ 0.006  & 0.814 $\pm$ 0.009  & 0.828 $\pm$ 0.008\\
    \hline
    \end{tabular}
\end{table}

\begin{table}[]
    \centering
    \begin{tabular}{clc}
    \hline
      & F1-score & Accuracy  \\
     \hline
    GAK & 0.571 $\pm$ 0.057 & 0.737 $\pm$ 0.040 \\
    \hline
    \end{tabular}
    \caption{Performance on the blood dataset.}
    \label{tab:VAR1}
\end{table}

\begin{figure}[!t]
    \centering
    \subfigure[TCK]
    {
        \includegraphics[trim = {42mm 93mm 42mm 57mm}, clip, width=0.32\linewidth]{SSI_TCKu.pdf}
    }%
        \subfigure[ssTCK]
    {
        \includegraphics[trim = {42mm 93mm 42mm 57mm}, clip, width=0.32\linewidth]{SSI_TCKss.pdf}
    }%
    \subfigure[sTCK]
    {
        \includegraphics[trim = {42mm 93mm 42mm 57mm}, clip, width=0.32\linewidth]{SSI_TCKs.pdf}
    }
    \subfigure[TCK$_{MNAR}$]
    {
        \includegraphics[trim = {42mm 93mm 42mm 57mm}, clip, width=0.32\linewidth]{SSI_TCK_mnar_u.pdf}
    }%
    \subfigure[ssTCK$_{MNAR}$]
    {
        \includegraphics[trim = {42mm 93mm 42mm 57mm}, clip, width=0.32\linewidth]{SSI_TCK_mnar_ss.pdf}
    }%
    \subfigure[sTCK$_{MNAR}$]
    {
        \includegraphics[trim = {42mm 93mm 42mm 57mm}, clip, width=0.32\linewidth]{SSI_TCK_mnar_s.pdf}
    }%
    \caption{Plot of the two-dimensional KPCA representation of the blood data obtained using the TCK and TCK$_{MNAR}$, respectively. The datapoints are color-coded according to whether the patients have SSI or not.}
    \label{fig: clustering results}
\end{figure}
}\fi

\section{Conclusions and future directions}
\label{sec: conclusion}
In this work, we presented  robust multivariate time series kernels capable of exploiting informative missing patterns and incomplete label information. 
In contrast to other frameworks that exploit informative missingness~\cite{DBLP:journals/corr/ChePCSL16, pmlr-v56-Lipton16}, which need complete label information, the time series cluster kernels are specially designed for situations in which no labels or only a few labels are available.
Lack of labels and large amounts of missing data are two challenges that characterize the medical domain, and therefore, we think the proposed kernels will be particularly useful in this domain, which we also demonstrated in this work through a case study of postoperative infections among colon rectal cancer patients. However, the kernels are not limited to this domain. We believe that these kernels could be useful tools in other application domains facing similar challenges.

A limitation of TCK$_{IM}$ is that if the missingness is by no means correlated with the outcome of interest, there will be limited gain in performance compared to the TCK, or might even a decrease in performance. For this reason it is important that the user has some domain knowledge and has some understanding about the process that led to missing values in the data, as illustrated in our case study from healthcare. 

An other limitation of the time series cluster kernels is that they are designed for MTS of the same length. A possible next step would be to work on a formulation that can deal with varying length.
In further work, we would also like to investigate the possibility of introducing a Bayesian formulation for the discrete modality in the mixed mode mixture models by putting informative priors over the parameters in the Bernoulli part of the model.



\section*{Conflict of interest}
The authors have no conflict of interest related to this work.

\section*{Acknowledgement}
 This work was partially funded by the Norwegian Research Council FRIPRO grant no. 239844 on developing the \emph{Next Generation Learning Machines}.
Cristina Soguero-Ruiz is partially supported by project TEC2016-75361-R from Spanish Government and by project DTS17/00158 from Institute of Health Carlos III (Spain).

The authors would like to thank Kristian Hindberg from UiT The Arctic University of Norway for his assistance on preprocessing and extracting the data from the EHR system. We would also like to thank Rolv-Ole Lindsetmo and Knut Magne Augestad from the University Hospital of North Norway, Fred Godtliebsen from UiT, together with Stein Olav Skrøvseth from the Norwegian Centre for E-health Research for helpful discussions throughout the study and manuscript preparation.




\if
In the LPS a time series is represented as a matrix of segments. Randomness is injected to the learning process by randomly choosing time segment (column in the matrix) and lag $p$ for each tree in the Random Forest.
For each tree a Bag-of-Words type compressed representation is created from the output of the leaf-nodes. The
final time series representation is created by concatenating the representation obtained from the individual trees and is  in turn used to design the similarity using a histogram intersection kernel. 

Given two multivariate time series $X^{(n)}$ and $X^{(m)}$, a formal expression for the LPS-kernel is 
\begin{equation} \label{eq: LPS}
    K(X^{(n)},X^{(m)}) = \frac{1}{R J} \sum\limits_{k=1}^{R J} \min (h^n_k, h^m_k),
\end{equation}
where $h^n_k$ is the $k$th entry of the concatenated Bag-of-Words representation $H(X^{(n)})$.  More precisely, $H(X^{(n)})$ is a concatenation of $R$-dimensional frequency vectors of instances in the terminal nodes from all trees.

Each tree generates a representation and the
final time series representation is obtained via concatenation. For simplicity, assume
that all trees contain the same number of terminal nodes R. The general case is easily
handled. Let H j (x n ) denote the R-dimensional frequency vector of instances in the
terminal nodes from tree g j for time series x n . We concatenate the frequency vectors
over the trees to obtain the final representation of each time series, denoted as H (x n ),
of length R  J (and modified obviously for non-constant R). Our representation
summarizes the patterns in the time series based on the terminal node distribution of
the instances over the trees.
\fi

\if
For each location, the subseries in the original data are concatenated to form a
new attribute. The internal model selects a random attribute as the response
variable then constructs a regression tree. A collection of these regression trees
are processed to form a new set of instances based on the counts of the number

of subseries at each leaf node of each tree. Algorithm 8 describes the process.
LPS can be summarised as follows:
Stage 1: Construct an ensemble of r regression trees.
1. Randomly select a segment length l (line 3)
2. Select w segments of length l from each series storing the locations in matrix
A (line 4).
3. Select w segments of length l from each di↵erence series storing the locations
in matrix B (line 5).
4. Generate the n · l cases each with 2w attributes (line 6).
5. Choose a random column from W as the response variable then build a
random regression tree (i.e. a tree that only considers one randomly selected
attribute at each level) with maximum depth of d (line 7).
Stage 2: Form a count distribution over each tree’s leaf node.
1. For each case x in the original data, get the number of rows of W that
reside in each leaf node for all cases originating from x.
2. Concatenate these counts to form a new instance. Thus if every tree had t
terminal nodes, the new case would have r · t features. In reality, each tree
will have a di↵erent number of terminal nodes.
Classification of new cases is based on a 1-nearest neighbour classification on
these concatenated leaf node counts.
\fi

\section*{References}
\bibliographystyle{elsarticle-num}
\bibliography{bibliography}

\end{document}